%% file: MAIN.tex
\documentclass[twoside]{zHenriquesLab-StyleBioRxiv}

\usepackage{blindtext}  
\usepackage{makecell}   
\usepackage{multirow}   
\usepackage{booktabs}   
\usepackage{cleveref}   
\usepackage{siunitx}    
\usepackage{threeparttable}
\usepackage{algorithm}       
\usepackage{algpseudocode}%
\usepackage{stfloats}
\usepackage{placeins}

\crefname{figure}{Fig.}{Figs.}
\Crefname{figure}{Figure}{Figures}
\crefname{table}{Table}{Tables}
\Crefname{table}{Table}{Tables}
\crefname{section}{Sec.}{Secs.}
\Crefname{section}{Section}{Sections}
\crefname{equation}{Eq.}{Eqs.}
\Crefname{equation}{Equation.}{Equations}

\newcommand{\dd}{\mathrm{d}} 
\newcommand{\bv}[1]{\mathbf{#1}} 

\setcitestyle{square}
\renewcommand{\thesubsection}{\thesection.\Alph{subsection}}
\DeclareCaptionFormat{smallformat}{\normalfont\sffamily\fontsize{9}{11}\selectfont#1#2#3}
\renewcommand\Affilfont{\color{color0}\normalfont\sffamily\fontsize{9}{10}\selectfont}

\hypersetup{allcolors=black}
\def\abstract{%
  \setlength{\parindent}{0pt}
  \ifbfabstract\small\bfseries\else\footnotesize\fi}

\makeatletter
\usepackage{fancyhdr}
\newcommand*\BioFooterLeft{%
  \footerfont
  \@leadauthor\ifnum\value{authors}>1\textit{\,et al.}\fi}
\newcommand*\BioFooterRight{%
  \footerfont arXiv\hspace{7pt}|\hspace{7pt}\thepage}

\fancypagestyle{plain}{\fancyhf{}\fancyfoot[LE,LO]{\BioFooterLeft}%
                              \fancyfoot[RE,RO]{\BioFooterRight}}
\makeatother
\makeatletter
\fancypagestyle{plain}{%
  \fancyhf{}                              
  \fancyfoot[RO]{\aBioRXiv}               
  
  }
\makeatother

\leadauthor{Liu} 

\begin{document}
\title{Downscaling Extreme Precipitation with Wasserstein Regularized
Diffusion}
\shorttitle{}


\author[1]{Yuhao Liu}
\author[2,4]{James Doss-Gollin}
\author[3]{Qiushi Dai}
\author[1,4]{Ashok Veeraraghavan}
\author[1,4,\Letter]{Guha Balakrishnan}

\affil[1]{~Department of Electrical and Computer Engineering, Rice University}
\affil[2]{~Department of Civil and Environmental Engineering, Rice University}
\affil[3]{~Department of Computer Science, Rice University}
\affil[4]{~Ken Kennedy Institute, Rice University}
\affil[ ]{{\phantom{\textsuperscript{1}}}\texttt{\{yuhao.liu,jdossgollin,qd8,vashok,guha\}@rice.edu}}

\maketitle

\begin{abstract}
\noindent \textit{Abstract} -- Understanding the risks posed by extreme rainfall events requires analysis of precipitation fields with high resolution (to assess localized hazards) and extensive historical coverage (to capture sufficient examples of rare occurrences). Radar and mesonet networks provide precipitation fields at 1 km resolution but with limited historical and geographical coverage, while gauge-based records and reanalysis products cover decades of time on a global scale, but only at 30–50 km resolution. To help provide high-resolution precipitation estimates over long time scales, this study presents Wasserstein Regularized Diffusion (WassDiff), a diffusion framework to downscale (super-resolve) precipitation fields from low-resolution gauge and reanalysis products. Crucially, unlike related deep generative models, WassDiff integrates a Wasserstein distribution-matching regularizer to the denoising process to reduce empirical biases at extreme intensities. 
Comprehensive evaluations demonstrate that WassDiff quantitatively outperforms existing state-of-the-art generative downscaling methods at recovering extreme weather phenomena such as tropical storms and cold fronts. 
Case studies further qualitatively demonstrate WassDiff's ability to reproduce realistic fine-scale weather structures and accurate peak intensities. 
By unlocking decades of high-resolution rainfall information from globally available coarse records, WassDiff offers a practical pathway toward more accurate flood-risk assessments and climate-adaptation planning.
\end{abstract}



\begin{corrauthor}
\texttt{guha@rice.edu} \\
\end{corrauthor}

\input{sections/intro.tex}

\input{sections/related_works.tex}
\input{sections/method.tex}
\input{sections/result.tex}
\input{sections/discussion.tex}

\begin{acknowledgements}
The authors gratefully acknowledge the support of this research by the National Science Foundation (NSF) award number ISS-2107313. Any opinions, findings, conclusions, or recommendations expressed in this paper are those of the authors and do not necessarily reflect the views of the sponsors.
\end{acknowledgements}

\section*{Bibliography}
\bibliography{ref_manual,ref_zotero}

\clearpage
\section*{\protect\centering Supplementary Material}
\addcontentsline{toc}{section}{Supplementary Material}
\input{sections/s_wasserstein.tex}
\input{sections/s_setup.tex}
\input{sections/s_baselines.tex}

\input{sections/s_metrics.tex}

\input{sections/s_ablation_study.tex}

\input{sections/s_tiled_diffusion.tex}


\end{document}

%% file: sections/intro.tex
\section{Introduction} \label{sec_intro}

\begin{figure}[t]
	\centering
	\includegraphics[width=\linewidth]{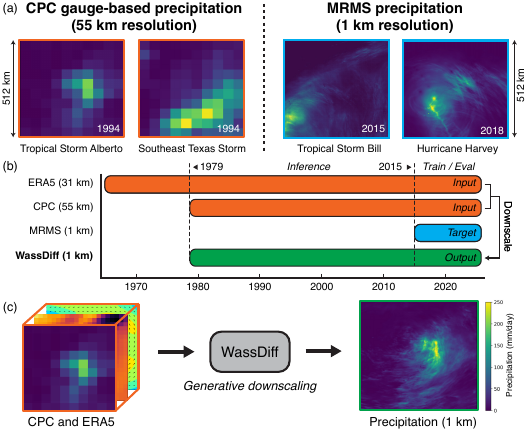}
	\caption{\textbf{Overview of precipitation signals used for downscaling in this study.}
	\textbf{(a)} Weather events recorded using sparse gauge instruments (CPC at \SI{55}{\kilo\meter} resolution) lack local precipitation dynamics, unlike modern radar that provides high resolution measurements (MRMS at \SI{1}{\kilo\meter} resolution).
	\textbf{(b)} Predating modern radar observations, there are decades of historical weather events documented only through coarse CPC gauge instruments and ERA5 reanalysis. In this study, we aim to downscale low-resolution historical weather events (1979 -- 2015) by leveraging high-resolution target training data (2015 -- present). \textbf{(c)}
    We propose a novel diffusion model named WassDiff to achieve this. WassDiff can downscale \SI{55}{\kilo\meter} inputs to \SI{1}{\kilo\meter} precipitation fields, recovering local and extreme precipitation dynamics.}
	\label{fig_overview}
\end{figure}

Precipitation dynamics profoundly affect Earth’s systems and human societies \citep{wright2019us, calvin2023ipcc, seneviratne2021weather}, and understanding these dynamics is essential to advance climate and hydrological science \citep{skofronick2017global} and to support critical applications such as stormwater management, floodplain mapping, flood insurance pricing, and hazard monitoring \citep{rozer2019probabilistic, sampson2015high, schneider2014gpcc, ahmed2021improved}. \emph{Extreme precipitation events}, such as convective storms and tropical cyclones, are particularly important to study because they drive most hydrologic risks and infrastructure failures \citep{hwang2024increasing}. 
However, capturing sufficient examples of such rare events requires decades-long observational archives.
In addition, these records should also ideally have high resolution ($\leq$ \SI{1}{\kilo\meter} ground sampling distance) to enable localizing high-impact effects \citep{fowler2021anthropogenic}.

Unfortunately, no existing precipitation data source provides both long-duration and high-resolution. On the one hand, analysis products derived from rain gauge records provide gap-free, accurate, and spatially consistent rainfall estimates over centuries \citep{lanza2008certified}. However, due to the sparsity of the gauges, these gridded products are only interpolated to coarse spatial resolution \citep{xie2007gaugebased}.
For example, Climate Prediction Center (CPC) Unified Precipitation~\citep{xie2007gaugebased} has only \SI{55}{\kilo\meter} resolution (\SI{0.5}{\degree} grid, see \cref{fig_overview}a). 
On the other hand, modern weather radar products~\citep{metoffice2003km} such as Multi-Radar Multi-Sensor (MRMS)~\citep{zhang2016multiradar} deliver high-resolution fields, but are only available over the last two decades (\cref{fig_overview}b). 
Hence, existing data sources pose a fundamental challenge: \emph{high-resolution precipitation records are typically short in duration, while long-term records typically suffer from low spatiotemporal resolution.}

A logical approach to tackle this challenge is to \textit{downscale} (or super-resolve) long-duration, low-resolution datasets using patterns learned from recent high-resolution data such as MRMS. Traditional dynamical \citep{dowell2022high, routray2010simulation} and statistical \citep{wilby1998statistical} downsampling variants are computationally expensive and/or limited in accuracy \citep{nishant2023comparison}. Machine learning (ML) downscaling methods offer a compelling alternative by leveraging advanced, data-driven algorithms to model complex atmospheric dynamics~\citep{veillette2020sevir, rampal2022highresolution, rodrigues2018deepdownscale}. In particular, modern deep generative models driven by neural networks---such as generative adversarial networks (GANs)~\citep{price2022increasing, leinonen2021stochastic,harris2022generative} and recent diffusion models~\citep{addison2022machine, mardani2025residual}---are particularly attractive due to their complex function capacities and ability to describe the inherent stochasticity in empirical distributions via Monte Carlo sampling.  
However, despite considerable advances in ML-based downscaling, accurate recovery of rainfall \emph{extremes} remains a challenge. By definition, extreme events reside in distribution tails, which are the most challenging regions to accurately model due to their low probabilities. Indeed, deep generative models are known to poorly capture rare events at distribution tails and instead over-represent distribution modes~\citep{bau2019seeinggangenerate,pandey2025heavytailed}. Our empirical evaluations demonstrate that existing generative downscaling approaches tend to produce incorrect \emph{lower} (diluted) precipitation estimates at weather extremes for this reason. 

This phenomenon is further exacerbated when downsampling from extremely coarse inputs, such as CPC gauge data. CPC observations are two to three times coarser than typical global weather forecasts \citep{gefs_2025,ifs_2025} and can have measurement errors such as those caused by wind-induced undercatch during high rainfall events \citep{pollock2018quantifying}. However, gauge‐based records provide decades of uninterrupted, ground-truth precipitation measurements, making them indispensable for long-term climate and hydrology research, and downscaling these products could enable a multitude of operational studies. Unfortunately, we find that recent state-of-the-art models designed to downscale global weather forecasts fare poorly in reconstructing extreme events when applied directly to coarse-scale gauge records~\citep{mardani2025residual,price2022increasing}. Hence, there is a need for novel methods that can accurately downscale gauge data to reliably recover a distribution over local precipitation dynamics.

To address these challenges, we introduce \emph{Wasserstein Regularized Diffusion} (WassDiff), a novel generative framework for downscaling precipitation estimates. WassDiff is a diffusion model~\citep{song2020scorebased} that leverages a distribution-matching regularizer in its denoising process during its training phase to reduce empirical biases at extreme intensities. This regularizer imposes a Wasserstein distance penalty~\citep{villani2009optimal} between model prediction and ground-truth distributions throughout the generative denoising process to suppress intensity miscalibrations. Wasserstein distance is particularly effective at capturing distributional shifts in precipitation data while remaining sensitive to changes in distribution tails (corresponding to extremes). 

We trained and evaluated WassDiff models for the task of generating \SI{1}{\kilo\meter} precipitation estimates from coarse CPC gauge-based precipitation and ERA5 reanalysis variables (essential atmospheric and environmental contexts that are linked to precipitation dynamics \citep{ghil2020physics}). Comprehensive quantitative evaluations show that WassDiff outperforms existing state-of-the-art downscaling models, and produces rainfall patterns with well-calibrated intensity estimates with respect to ground-truth radar measurements. 
Qualitative case studies of extreme weather phenomena, like tropical storms and cold fronts, further demonstrate WassDiff's operational value by predicting appropriate spatial patterns while capturing extremes. In summary, this work makes several contributions: 
\begin{enumerate}
	\item A Wasserstein Regularized Diffusion framework that effectively reduces biases in the denoising process and enforces proper intensity calibration.
	\item A generative downscaling WassDiff diffusion model trained to reconstruct \SI{1}{\kilo\meter} precipitation from coarse CPC gauge-based precipitation (\SI{55}{\kilo\meter}) and ERA5 reanalysis data (\SI{31}{\kilo\meter}) over the contiguous United States.
    \item Extensive benchmarks and experiments demonstrating that WassDiff improves downscaling performance over existing baselines on extreme precipitation recovery, along with case studies showcasing robust downscaling performance across various weather phenomena, such as tropical storms and cold fronts.
    \item An open-source framework~\footnote{Source code will be available upon publication.} to generate extensive \SI{1}{\kilo\meter}-resolution precipitation estimates using gauge records and reanalysis that are available globally for decades.
\end{enumerate}

%% file: sections/related_works.tex
\section{Related works} \label{appendix_related}

\begin{figure*}[ht]
    \centering
    \includegraphics[width=1.0\linewidth]{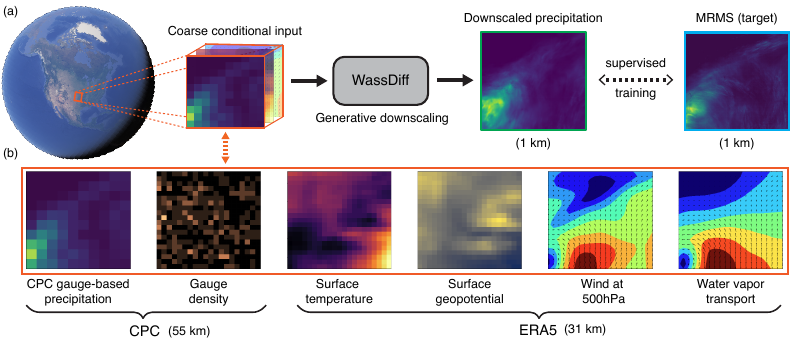}
    \caption{\textbf{Overview of the proposed downscaling model WassDiff.}
    (a) WassDiff generates \SI{1}{\kilo\meter} precipitation data conditioned on coarse-scale inputs.
    (b) Coarse-scale inputs are acquired from CPC gauge records (at \SI{55}{\kilo\meter}) and  ERA5 reanalysis (at \SI{31}{\kilo\meter}). These ERA5 reanalysis variables provide essential atmospheric and environmental context linked to precipitation dynamics.}
    \label{fig_method}
\end{figure*}

Early ML downscaling approaches often employ Convolutional Neural Networks (CNNs) to translate coarse inputs to fine-scaled fields \citep{veillette2020sevir, rampal2022highresolution, rodrigues2018deepdownscale}.
These CNNs act as regression models trained to produce a single deterministic prediction—namely, the mode of the distribution---yielding blurry outputs without small‐scale rainfall structures \citep{ravuri2021skillful}. 
However, since atmospheric processes at km-scale are inherently stochastic \citep{selz2015upscale}, it is more appropriate to adopt generative downscaling models that learn to produce full probabilistic distributions rather than a single point estimate.

Initially, Generative Adversarial Networks (GANs) have been used in downscaling precipitation \citep{leinonen2021stochastic, price2022increasing, vosper2023deep}.
Both \citet{price2022increasing} and \citet{harris2022generative} demonstrated downscaling global forecasts to match radar precipitation measurements, from $32$ to \SI{4}{\kilo\meter} resolution and from $10$ to \SI{1}{\kilo\meter} resolution, respectively. 
Generally speaking, training GANs poses several challenges, including mode collapse, training instabilities, and difficulty capturing long tails of the distributions \citep{xiao2022tackling, kodali2017convergence, salimans2016improved}.

Recently, diffusion models have been introduced as an alternative to GANs for their sample diversity and training stability \citep{ho2020denoising, dhariwal2021diffusion}.
\citet{addison2022machine} used a score-based diffusion model \citep{song2020scorebased} to synthesize rainfall at \SI{8.8}{\kilo\meter}, downscaled from a general circulation model at \SI{64}{\kilo\meter}.
The previous state-of-the-art generative downscaling approach CorrDiff \citep{mardani2025residual} uses a two-step process---using a regression to predict mean and diffusion to predict variance.
CorrDiff can synthesize consistent radar reflectivity patterns, particularly in coherent weather structures like typhoons.
However, radar reflectivity---known to exhibit biases, such as signal attenuation in heavy rain---does not directly translate to accurate rain rate estimates, which limits the model's utility to downstream risk assessment tasks. 

Downscaling studies also differ in operational utility based on their data domain, which generally falls into three categories: synthetic, numerical forecasts, and observation records.
Early pioneering works~\citep{leinonen2021stochastic, vosper2023deep} experiment with \textit{synthetic} data pairs, aiming at recovering high-resolution rainfall from artificially blurred (downsampled) versions of the same fields. 
Such ``deblurring step''---inverting a fixed smoothing operator---requires no bias or noise correction and is considered relatively trivial with limited operational value.
In contrast, recent efforts~\citep{mardani2025residual,price2022increasing,harris2022generative,addison2022machine,rampal2022highresolution,rodrigues2018deepdownscale} focus on \textit{numerical forecast} data \citep{ifs_2025,gefs_2025}, downscaling from coarse-scale global forecasts to high-resolution regional forecasts.
Global and regional forecasts share a high degree of physical consistency~\citep{giorgi2019thirty} but differ in grid resolution, motivating the use of generative downscaling to add realistic local weather patterns.
In the context of forecasting and nowcasting, these generative downscaling studies enter as a cost-effective alternative to regional forecast models that rely on numerical physics.

This work operates in the third data category: \textit{historical observation records}, which poses unique challenges.
Unlike synthetic data or even forecasts, gauge records~\cite{xie2007gaugebased} also contain measurement errors, such as wind-induced undercatch during high rainfall events \citep{pollock2018quantifying}.
Gridded gauge records (\SI{55}{\kilo\meter}) are also coarser than typical weather forecast models (e.g., \SI{10}{\kilo\meter} for IFS \citep{ifs_2025} and \SI{25}{\kilo\meter} for GEFS \cite{gefs_2025}).
The downscaling ratio in our current work ($55\times$) is higher than any prior generative downscaling work, adding downscaling uncertainty.
Since our goal here is to accurately recover extreme precipitation---a particularly challenging task when dealing with very coarse grid cells---addressing these limitations demands an innovative approach that effectively recovers the extreme and local precipitation phenomena that are absent in the coarse measurements. 
For completeness, we empirically compare WassDiff against prior work, demonstrating its effectiveness in accurately reconstructing extreme rainfall events.

%% file: sections/method.tex
\section{Methods} \label{sec_method}

\subsection{Multivaraite precipitation downscaling}

Within a geographical region bounded by some coordinates, we extract coarse-scale inputs from two sources, as seen in \cref{fig_method}.
We first extract CPC Global Unified Gauge-based Precipitation \citep{xie2007gaugebased} ($\mathbf{y}_p\in\mathbb{R}^{m' \times n'}$) at \SI{55}{\kilo\meter} resolution, along with the corresponding gauge density ($\mathbf{y}_d \in \mathbb{R}^{m' \times n'}$). 
We additionally extract six ERA5 \citep{era5data} reanalysis variables ($\mathbf{y}_{era5} \in \mathbb{R}^{m'' \times n'' \times c_{era5}}$) from the same region at \SI{31}{\kilo\meter} resolution.
ERA5 variables include surface temperature, surface geopotential (i.e., elevation), wind at \SI{500}{\hecto\pascal} ($\mu$ \& $\nu$ components), and vertical integral of water vapor transport ($n$ and $e$ components).
These ERA5 variables provide essential atmospheric and environmental context linked to precipitation dynamics \citep{back2005relationship,ghil2020physics,li2024revealing,ogorman2015precipitation}. 
We bilinearly upsample all conditional inputs to target resolution (\SI{1}{\kilo\meter}), yielding a multivariate tensor $\mathbf{y} = [f_{\uparrow}(\mathbf{y}_p), f_{\uparrow}(\mathbf{y}_{era5}), f_{\uparrow}(\mathbf{y}_d$)], where  $\mathbf{y} \in \mathbb{R}^{m \times n \times c_{in}}$, and $f_{\uparrow}$ denote bilinear upsampling.
We use \SI{1}{\kilo\meter} precipitation from Multi-Radar Multi-Sensor (MRMS) \citep{zhang2016multiradar} as ground truth.
These data collectively form a training set comprising a large corpus (5.5 years) of precipitation events sampled in the CONUS region from Sept. 4, 2016 -- Dec 31, 2020. 
See Supplementary for additional details.

We cast our downscaling task as follows: given a set of low-resolution data---gauge-based precipitation, gauge density, and six ERA5 variables---we generate high-resolution precipitation estimates $\bv{x}$.
More precisely, we aim to model the probability density function $p(\mathbf{x} | \mathbf{y})$ through ensemble inference.

\subsection{Generative downscaling via diffusion models} \label{sec_background}

Diffusion models \citep{song2020scorebased} can learn the conditional data distribution $p(\mathbf{x|\bv{y}})$ using a neural network to reverse a predefined forward corruptive noising process.
Following prior work~\citep{song2020scorebased}, we formulate the forward and reverse diffusion process using stochastic differential equations (SDEs)~\citep{song2020generative}.
Consider $p_{data}$ as the true data (i.e., target) distribution and $p_T$ as the prior distribution (i.e., noise).
The SDE for the forward diffusion process $\{\mathbf{x}(t)\}^T_{t=0}$ indexed by a continuous time variable $t \in [0, T]$ is described as
\begin{equation} \label{eq_forward_sde}
	\dd \mathbf{x} = \mathbf{f}(\mathbf{x}, t) \dd t + g(t) \dd\mathbf{w},
\end{equation} 
where $\mathbf{w}$ is the standard Wiener process (or Brownian motion), $\mathbf{f}(\cdot{}, t): \mathbb{R}^d \rightarrow \mathbb{R}^d$ is the drift coefficient of $\mathbf{x}(t)$, and $g(\cdot): \mathbb{R} \rightarrow \mathbb{R}$ is the diffusion coefficient of $\mathbf{x}(t)$. To generate data samples $\mathbf{x}(0) \sim p_0(\bv{x}|\bv{y})$, we start by sampling the prior distribution $\mathbf{x}(T) \sim p_T$ and follow the reverse-time SDE:
\begin{equation} \label{eq:reserve_sd}
	\dd \mathbf{x} = [\mathbf{f}(\bv{x}, t) - g(t)^2 \nabla_{\bv{x}} \text{log}p_t(\bv{x}|\bv{y})]\dd t + g(t) \dd \bar{\bv{w}},
\end{equation} 
where $\bar{\bv{w}}$ is the standard Weiner process when time flows backward from $T$ to $0$, and $\nabla_{\bv{x}} \text{log}p_t(\bv{x} | \bv{y})$ describe the conditional \textit{score} (i.e., the gradient of the log probability density w.r.t. data) at an intermediate time step $t$.
The ability to generate samples requires an accurate estimate of the true score function $\nabla_{\bv{x}} \text{log}p_t(\bv{x}| \bv{y})$.
This is achieved by training a time-dependent score-based model $\bv{s}_{\bv{\theta}}(\bv{x}, \bv{y},t)$ to approximate $\nabla_{\bv{x}} \text{log}p_t(\bv{x}|\bv{y})$ via the denoising score-matching \citep{song2020scorebased} training objective:
\begin{align} \label{eq_score_matching_conditional}
\boldsymbol{\theta^*} = ~& \underset{\boldsymbol{\theta}}{\arg\min}~\mathbb{E}_t \Big\{  \lambda(t) \mathbb{E}_{\bv{x}(0)}\mathbb{E}_{\bv{x}(t)|\bv{x}(0)} \big[ \|\bv{s}_{\bv{\theta}}(\bv{x}(t), \bv{y}, t) \nonumber \\
& - \nabla_{\bv{x}(t)} \log p_{0t}(\bv{x}(t) | \bv{x}(0))\|^2_2 \big]  \Big\},
\end{align}
where $\lambda: [0, T] \rightarrow \mathbb{R}_{>0}$ is a positive weighting function, $t$ is uniformly sampled over $[0, T]$, $\bv{x}(0) \sim p_0(\bv{x})$ and $\bv{x}(t) \sim p_{0t}(\bv{x}(t) | \bv{x}(0))$, where $p_{0t}(\bv{x}(t) | \bv{x}(0))$ is the transition kernel from $\bv{x}(0)$ to $\bv{x}(t)$.

With sufficient data and model capacity, in theory, score matching (\cref{eq_score_matching_conditional}) leads to an optimal solution  $\bv{s}_{\boldsymbol{\theta}^*}(\bv{x},\bv{y}, t)$ that is equal the true score function $\nabla_{\bv{x}} \text{log}p_t(\bv{x}|\bv{y})$ for almost all $\bv{x}$, $\bv{y}$ and $t$ \citep{song2020scorebased}.
However, in practice, obtaining such optimal $\bv{s}_{\boldsymbol{\theta}^*}(\bv{x},\bv{y}, t)$ can be challenging. 

\subsection{Wasserstein Regularized Diffusion} \label{sec_wass_dist}


While in theory, score-matching training objective (Eq.~(\ref{eq_score_matching_conditional})) can perfectly match the target distribution, in practice, however, we empirically found that standard score-based diffusion models tend to generate samples with inaccurate intensity estimates, which can lead to unacceptable errors.
For precipitation, intensity is defined as rain rate (\SI{}{\milli\meter\per\day}).

\begin{figure}[t]
	\centering
	\includegraphics[width=1.0\linewidth]{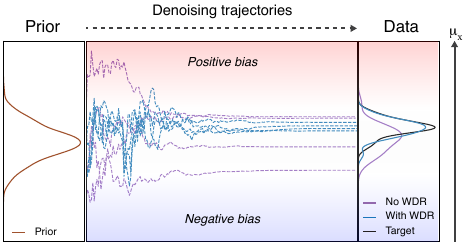}
	\caption{
	\textbf{Wasserstein Distance Regularization (WDR) mitigates biases during denoising.}
		With WDR, sample average intensity ($\mu_\bv{x}$) is well controlled in the denoising process (dashed blue lines), resulting in a sample intensity distribution (blue curve) that closely matches the target distribution (black curve).
		Conventional score-matching objective does not \textit{explicitly} control biases during denoising (dashed purple lines), and the resulting intensity distribution (purple curve) deviates from the target distribution.}
	\label{fig_sde}
\end{figure}

The data generation process for a standard score-based diffusion model is illustrated by the purple lines in \cref{fig_sde}.
All trajectories in \cref{fig_sde} are initialized with the same condition $\bv{y}$ but with different samples from the prior distribution $p_T$.
The purple dashed lines visualize the progression of average intensity ($\mu_\bv{x}$) during the denoising process.
Poor calibration produces denoising trajectories with large biases.
These trajectories ultimately lead to a sample distribution (solid purple line) that deviates from the target distribution (solid black line)---in this case, with significant negative bias.

We aim to mitigate biases by correcting deviations in the denoising trajectory.
Intuitively, we seek a mechanism to penalize intensity deviations \textit{at each denoising step}, since biases can accumulate early in the denoising process, as seen in \cref{fig_sde}.
Calculating such penalization requires a way to quantify the discrepancy---a distance metric between the sample and target distribution at each denoising step.
Accurate precipitation estimates demand a distance metric to be sensitive to changes in distribution tails (corresponding to extremes).
This criterion renders conventional distance metrics---such as the Jensen-Shannon divergence \citep{lin1991} and the Kullback–Leibler divergence \citep{kullback1951} that mainly capture differences in central modes---inadequate for this task.
See Supplementary for details.


We propose using the Wasserstein distance~\citep{villani2009optimal} to quantify distribution shifts during denoising.
The Wasserstein distance, which measures the minimal ``transport cost'' required to morph one probability distribution into another, exhibits stronger sensitivity to discrepancies in distribution tails. Since gridded precipitation fields inherently reside in high-dimensional spaces, we employ the \textit{sliced} Wasserstein distance \citep{bonneel2015sliced} as a tractable approximation. 
Sliced Wasserstein distance is efficient to compute and has been shown to be a statistically effective metric between two distributions in high-dimensional spaces, given a high number of projections~\cite{kolouri2019generalized, nguyen2024energy}. 
In practice, we use 100 projections, which provide sufficient context of an image's intensity profile with reasonable computational time. 
See Supplementary for algorithmic details.




\begin{figure*}[hb]
	\centering
	\includegraphics[width=1.0\linewidth]{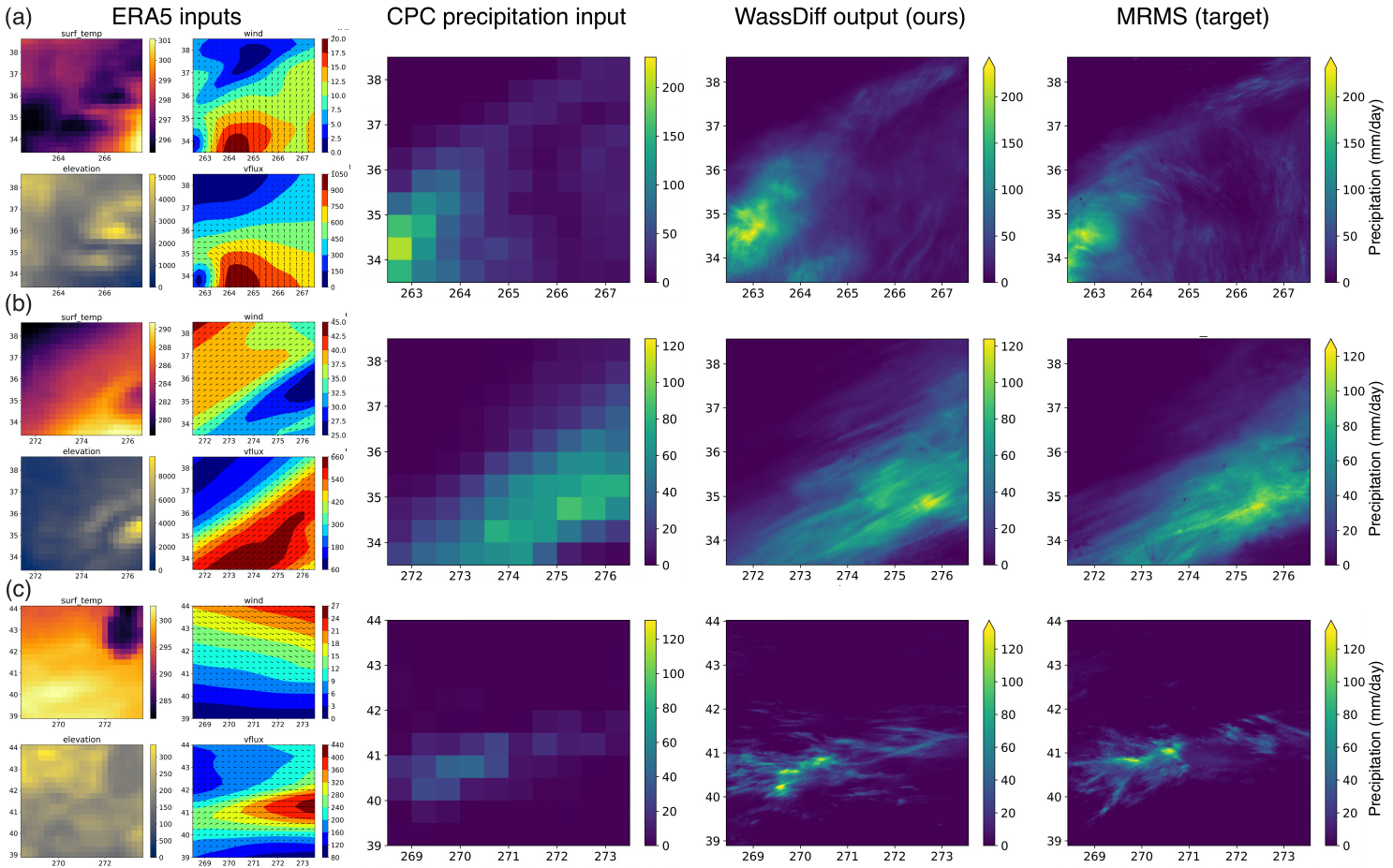}
	\caption{\textbf{Demonstration of precipitation downscaling of extreme weather events.}
        Conditioned on ERA5 inputs (\SI{31}{\kilo\meter}) and CPC precipitation (\SI{55}{\kilo\meter}), WassDiff produces \SI{1}{\kilo\meter} precipitation estimates, shown next to \SI{1}{\kilo\meter} MRMS targets. Downscaling is demonstrated on these weather events:
	(a) Tropical Storm Bill, 2015-06-18 UTC. (b) A cold front, 2015-12-02 UTC. (c) A hailstorm, 2015-06-11 UTC.
    WassDiff produces structured patterns---such as spiral bands in (a) and a sharp rain boundary in the lower right corner of (b)---consistent with the MRMS targets while accurately capturing extreme rainfall.}
	\label{figure_vis_results}
\end{figure*}

We modify the training objective of the standard score-based diffusion model (Eq.~(\ref{eq_score_matching_conditional})) to now incorporate sliced Wasserstein Distance $W^S$ between the partially denoised samples $\bv{x}(t)$ and target $\bv{x}$ at each step $t$:
\begin{align} \label{eq_score_emd}
\boldsymbol{\theta^*} = ~& \underset{\boldsymbol{\theta}}{\arg\min}~\mathbb{E}_t \Big\{ \lambda(t) \mathbb{E}_{\bv{x}(0)}\mathbb{E}_{\bv{x}(t)|\bv{x}(0)} \nonumber \\
& \big[ (1-\alpha) \underbrace{\|\bv{s}_{\bv{\theta}}(\bv{x}(t), \bv{y}, t) \nonumber - \nabla_{\bv{x}(t)} \log p_{0t}(\bv{x}(t) | \bv{x}(0))\|^2_2}_{\text{score-matching}} \nonumber \\
& + \alpha \underbrace{W^S(\mathbb{P}_{\bv{x}(0)}, \mathbb{P}_{\bv{x}})}_{\text{regularization}} \big] \Big\},
\end{align}
where $\alpha \in [0, 1]$ is a scalar coefficient that controls the strength of regularization.
Following a preliminary hyperparameter search, we set $\alpha=0.2$ in this work.
Since $W^S$ regulates the standard score-matching denoising trajectories, we term this mechanism \textit{Wasserstein Distance Regularization} (WDR), which underpins the foundation of a new class of diffusion models---\textit{Wasserstein Regularized Diffusion} (WassDiff).

Fig.~\ref{fig_sde} shows the effect of WDR during the denoising process.
Denoising trajectories regulated by WDR are indicated by dashed blue lines.
Under WDR, biases in the denoising process are well controlled, resulting in a sample distribution (solid blue line) closely matching with target distribution (solid black line).
Under the context of precipitation downscaling, WDR leads to accurate precipitation estimates that are consistent with the target distribution.

%% file: sections/result.tex
\section{Results} \label{sec_result}

\subsection{Case studies of extreme weather events}

\begin{figure*}[h]
	\centering
\includegraphics[width=1.0\linewidth]{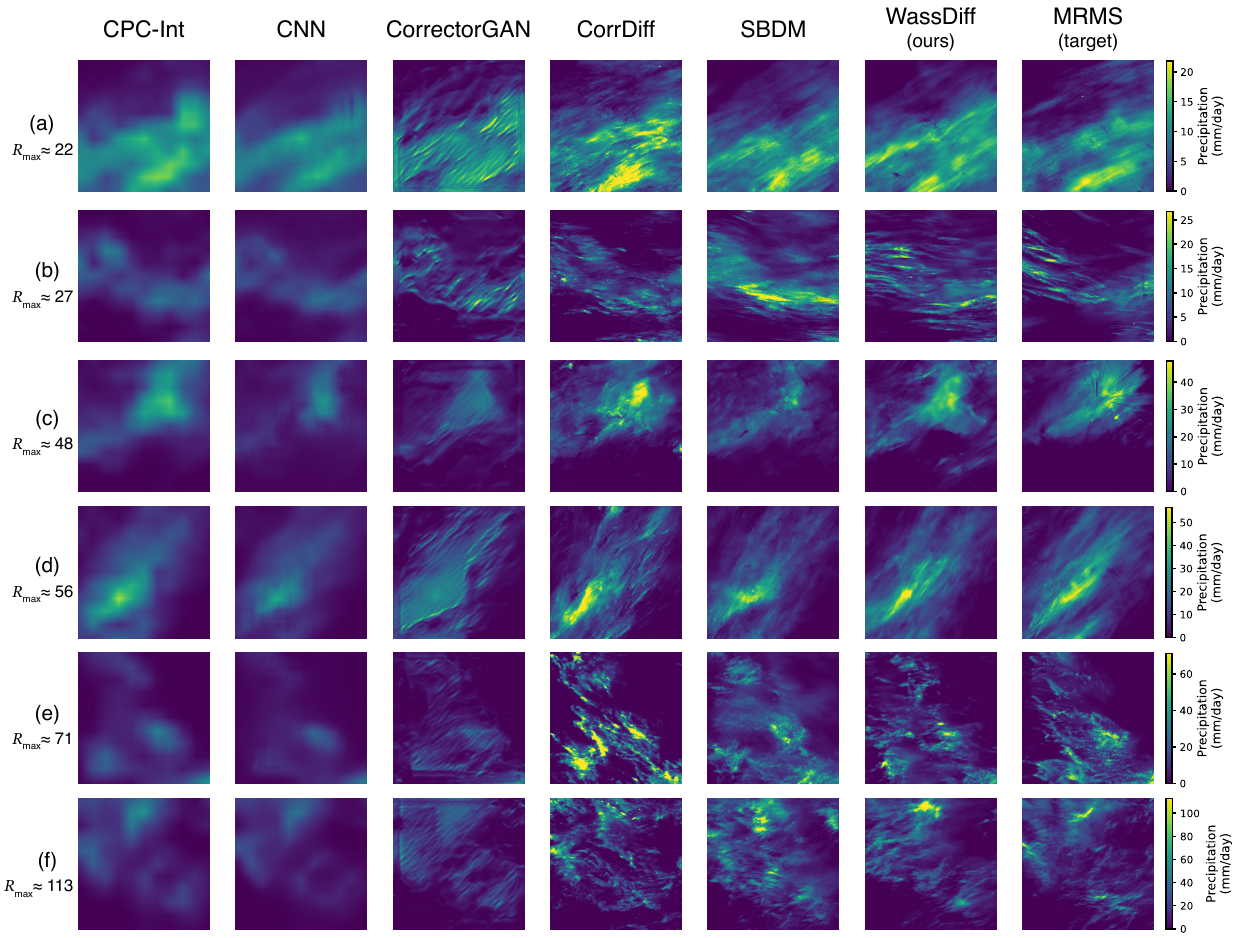}
	\caption{\textbf{Visual comparison of outputs from different downscaling methods.}
	Six representative examples from the test dataset are shown for each downscaling method in Table~\ref{tab:model_comparison}. 
	The samples are arranged in the order of increasing max rainfall intensity ($R_\text{max}$) to highlight performance differences.
	Deterministic methods (CPC-Int and CNN) yield blurry predictions that fail to capture extreme precipitation dynamics. 
	Among generative models, diffusion-based approaches (last three) demonstrate noticeably higher perceptual quality than CorrectorGAN.
	Our model (WassDiff) recovers extreme precipitation signals more reliably than the other two diffusion-based models (CorrDiff and SBDM).}
	\label{fig_additional_results}
\end{figure*}

Operational meteorologists value case studies, as aggregated skill scores can mask critical details of the downscaling performance.
In \cref{figure_vis_results}, we present three types of extreme weather events to further demonstrate reconstruction skills.
For each event, we include ERA5 inputs (at \SI{31}{\kilo\meter}), low-resolution CPC precipitation input (at \SI{55}{\kilo\meter}), WassDiff output (at \SI{1}{\kilo\meter}), and MRMS ground truth (at \SI{1}{\kilo\meter}), hereafter referred to as \textit{targets}.
ERA5 inputs shown \cref{figure_vis_results} are surface temperature, surface geopotential (i.e., elevation), wind at \SI{500}{\hecto\pascal}, and water vapor transport (their impact on downscaling performance is shown in Supplementary).
We use Universal Coordinated Time (UTC) for all date and time references in this paper.

Fig.~\ref{figure_vis_results}(a) shows reconstruction results for Tropical Storm Bill (2015), a large-scale coherent structure.
While the coherent structures (such as spiral bands of clouds emanating from the storm center) are entirely missing from coarse CPC input, our model produces those patterns akin to the MRMS target by leveraging ERA5 ancillary variables.
This is a reassuring sign that our diffusion model produces output reminiscent of the appropriate multivariable physics between precipitation and other climate variables, such as wind and temperature.

Fig.~\ref{figure_vis_results}(b) presents a cold front, which is a frontal system
with a sharp boundary in the atmosphere, where a colder air mass displaces a warmer air mass in the upward direction.
Upward displacement of warm air leads to cooling, followed by condensation and, ultimately, rainfall.
Downscaling frontal systems provides utility because intense rainfall tends to occur near the frontal boundary, which is captured by our diffusion output and MRMS but absent in the coarse CPC input.
The magnitude of heavy rainfall (lower right corner) is well-calibrated to the MRMS target.

Fig.~\ref{figure_vis_results}(c) shows a hailstorm---observed near Minooka, IL---
a form of solid precipitation associated with strong thunderstorms with intense updrafts that carry water droplets into extremely cold parts of the atmosphere, causing them to freeze and ultimately resulting in fallen ice crystals (i.e., hailstones).
Ice crystals can be resolved by weather radars like MRMS but not gauge-based measurements like CPC, as seen in Fig.~\ref{figure_vis_results}(c).
WassDiff captures such isolated, localized precipitation with a well-calibrated intensity consistent with the target.

\subsection{Qualitative comparison of downscaling models} \label{sec_vis_compare}

In this section, we visually compared WassDiff's downscaling performance against five baselines: (1) bilinearly-interpolated CPC precipitation (CPC-Int), (2) a convolutional neural network (CNN)~\citep{veillette2020sevir}, (3) a bias-correcting Generative Adversarial Network (CorrectorGAN)~\citep{price2022increasing}, (4) a recent state-of-the-art diffusion generative downscaling model (CorrDiff)~\citep{mardani2025residual}, and (5) a standard score-based diffusion model (SBDM), which uses traditional score-matching denoising objective~\citep{song2020scorebased}.
CPC-Int and CNN are deterministic models, while the rest are generative models capable of performing ensemble (probabilistic) forecasting.
Among them, CorrDiff, SBDM, and WassDiff are diffusion-based models.
All models are trained from scratch on the same dataset and receive identical input (CPC gauge-based precipitation + ERA5), except for CPC-Int, which is only conditioned on CPC and requires no training.
See Appendix Supplementary for experimental setups and details on baseline models.

\cref{fig_additional_results} visualizes outputs from these six downscaling methods along with \SI{1}{\kilo\meter} MRMS targets.
We present a variety of weather events arranged in increasing max rainfall intensity ($R_\text{max}$).
Notably, deterministic methods (CPC-Int and CNN) yield blurry predictions that fail to capture extreme precipitation values---a limitation especially evident in heavy-to-extreme rainfall cases (rows e and f).
Unlike deterministic models that estimate distributional modes, generative models sample from the full probability distribution, producing more realistic texture details depending on their respective architecture.
In particular, CorrectorGAN produces local textures; however, its outputs often feature wavy patterns that deviate from true precipitation dynamics while systematically underestimating the extremes. 
Meanwhile, the three diffusion-based models (CorrDiff, SBDM, and WassDiff) deliver noticeably higher perceptual quality with patterns closely resembling the observed targets.
The perceptual differences among the three diffusion models are subtle, with CorrDiff producing the sharpest textures and WassDiff most closely matching true radar observations.
However, these diffusion models differ in their predicted intensity.
Unfortunately, CorrDiff tends to struggle at extreme values (under-estimated in rows b and f, while over-estimated in a, c, and e).
Similarly, SBDM also inaccurately estimates extreme precipitation values (rows b and c).
WassDiff, however, produces intensity estimates that better align with the targets, while effectively reconstructing heavy-to-extreme rainfall.

\subsection{Quantitative reconstruction skills}

\begin{table*}[t]
\centering
\caption{
\textbf{Quantitative evaluation of downscaling six methods across 442 test samples.}
This table includes two deterministic models (CPC-Int and CNN \citep{veillette2020sevir}) and four generative models (CorrectorGAN \citep{price2022increasing}, CorrDiff \citep{mardani2025residual}, SBDM, and WassDiff).
Each generative model has 13 ensemble members.
Metrics include Mean Squared Error (MAE), Critical Success Index (CSI), bias, Continuous Ranked Probability Score (CRPS), Heavy Rain Region Error (HRRE), Mesoscale Peak Precipitation Error (MPPE), and LPIPS for perceptual differences.
We highlight the best performance in \textbf{bold} and the second-best in \underline{underline}.
}
\label{tab:model_comparison}
\begin{tabular}{@{}lccccccccc@{}}
\toprule
Model & MAE~$\downarrow$ & CRPS~$\downarrow$ & CSI~$\uparrow$ & Bias & HRRE $\downarrow$ & MPPE $\downarrow$ & LPIPS $\downarrow$ \\ \midrule
CPC-Int & 2.54 {\scriptsize $\pm$ 2.00}& - & 0.31 {\scriptsize $\pm$ 0.30} & \underline{-0.19} {\scriptsize $\pm$ 1.56} & 952 {\scriptsize $\pm$ 2661}& 22.04 {\scriptsize $\pm$ 23.50} & 0.55 {\scriptsize $\pm$ 0.21}\\
CNN & \underline{2.50} {\scriptsize $\pm$ 2.09}& -& 0.23 {\scriptsize $\pm$ 0.29} & -1.14 {\scriptsize $\pm$ 1.89} & 1269 {\scriptsize $\pm$ 3608}& 30.00 {\scriptsize $\pm$ 48.38} & 0.56 {\scriptsize $\pm$ 0.20} \\
CorrectorGAN & 2.65 {\scriptsize $\pm$ 2.25}& 2.50 {\scriptsize $\pm$ 2.13}& 0.28 {\scriptsize $\pm$ 0.28}& -1.07 {\scriptsize $\pm$ 1.77}& 1283 {\scriptsize $\pm$ 3834} & 20.30 {\scriptsize $\pm$ 23.85}& 0.48 {\scriptsize $\pm$ 0.11}\\
CorrDiff & \underline{2.50} {\scriptsize $\pm$ 1.96}& \textbf{1.78} {\scriptsize $\pm$ 1.54}& \textbf{0.33} {\scriptsize $\pm$ 0.29}& -0.20 {\scriptsize $\pm$ 1.45}& \underline{848} {\scriptsize $\pm$ 2301} & \underline{14.30} {\scriptsize $\pm$ 14.79}& \textbf{0.41} {\scriptsize $\pm$ 0.09}\\
SBDM & 2.75 {\scriptsize $\pm$ 2.49}& 2.04 {\scriptsize $\pm$ 2.04}& 0.24 {\scriptsize $\pm$ 0.27}& -1.24 {\scriptsize $\pm$ 2.55}& 978 {\scriptsize $\pm$ 2847} & 16.72 {\scriptsize $\pm$ 17.69}& 0.44 {\scriptsize $\pm$ 0.11}\\
\textbf{WassDiff} & \textbf{2.48} {\scriptsize $\pm$ 2.10}& \underline{1.85} {\scriptsize $\pm$ 1.55}& \underline{0.32} {\scriptsize $\pm$ 0.30}& \textbf{-0.17} {\scriptsize $\pm$ 1.30}& \textbf{682} {\scriptsize $\pm$ 2142}& \textbf{12.78} {\scriptsize $\pm$ 15.54}& \underline{0.42} {\scriptsize $\pm$ 0.14}\\
\bottomrule
\end{tabular}
\end{table*}

We continue downscaling evaluation by benchmarking downscaling performance against \SI{1}{\kilo\meter} MRMS ground truth targets.
Table~\ref{tab:model_comparison} shows the skill scores across 442 test samples randomly selected in the contiguous United States (CONUS) from May 8, 2015 -- Sept 3, 2016, a temporal range unseen during training.
We use seven metrics to highlight different aspects of model performance.
Mean absolute error (MAE) quantifies the average error between the ensemble mean and the target, while Continuous Ranked Probability Score (CRPS) compares the predicted probabilities to the target.
Critical Success Index (CSI) reflects the categorical forecast performance at the threshold \SI{10}{\milli\meter\per\day}.
Heavy Rain Region Error (HRRE) \citep{chen2022rainnet} and Mesoscale Peak Precipitation Error (MPPE) \citep{chen2022rainnet} reflect model performance for heavy and extreme rainfall, respectively.
LPIPS \citep{zhang2018unreasonable} reflects the perceptual differences between model outputs and targets.
For each metric, we report the mean and the standard deviation.
Refer to the Supplementary for detailed metric definitions.

For deterministic skill metrics (MAE, CSI, and bias), the best-performing generative models only marginally improve upon the best deterministic models.
This is mainly due to the underlying CPC gauge readings (which all methods use as input) being, on average, highly accurate \citep{lanza2008certified}.
CNN, trained to minimize Mean Squared Error (MSE), marginally improves the MAE score compared to CPC-Int.
In contrast, generative models are trained to produce the entire probabilistic distribution and traditionally suffer from a slightly degraded MAE \citep{mardani2025residual}.
Nonetheless, WassDiff has the highest deterministic skills, demonstrating its effectiveness at predicting intensity values closely matching the corresponding targets, achieving the highest accuracy and lowest bias among all baselines.

Deterministic models tend to yield blurry predictions (shown in \cref{fig_additional_results}) that do not capture extreme rainfall events.
Consequently, HRRE and MPPE---which reward accurate heavy and extreme rainfall predictions---significantly favor generative models.
Diffusion models generally deliver superior performance compared to CorrectorGAN, which relies on an older Generative Adversarial Network (GAN) that is prone to training instability and mode collapse.
Notably, WassDiff achieves the highest HRRE and MPPE scores, significantly improved from the second-best model, CorrDiff ($+24\%$ HRRE and $+12\%$ MPPE).
This underscores WassDiff's effectiveness at recovering extreme precipitation signals.

WassDiff achieves the second-best CPRS, slightly falling behind ($-4\%$) the previous state-of-the-art baseline CorrDiff, which we suspect can be partially explained by the ``under-dispersive" nature of CorrDiff's two-step approach (UNet to predict mean, diffusion to predict variance) \citep{mardani2025residual}.
Having under-dispersive predictions may risk observations falling outside of the prediction spread, which has implications for extreme rainfall (discussed in \cref{subsec:qq}).
CorrDiff's two-step approach contrasts with WassDiff's full diffusion approach; the latter offers broader distributional coverage.
WassDiff and CorrDiff achieve the highest perceptual scores in LPIPS, consistent with their realistic textures in \cref{fig_additional_results}, with WassDiff being slightly worse than CorrDiff ($-2\%$), consistent with CorrDiff's slightly sharper outputs in \cref{fig_additional_results}.
Overall, WassDiff shows robust downscaling performance, yielding well-calibrated precipitation intensities while accurately recovering extreme precipitation, underscoring its advancement over existing methods.
See Supplementary for ablation studies.

\subsection{Quantile-quantile analysis} \label{subsec:qq}

\begin{figure*}[htb]
	\centering
	\includegraphics[width=\linewidth]{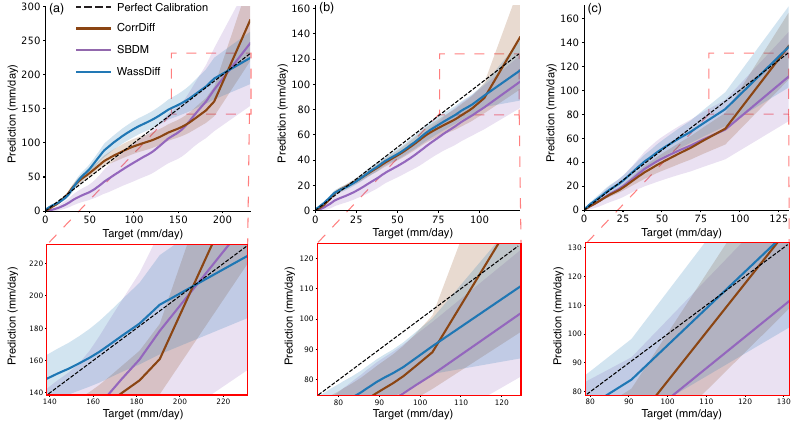}
	\caption{\textbf{WassDiff produces better-calibrated rainfall estimates than baseline generative downscaling models.}
	This figure shows quantile-quantile analysis for three extreme weather events referenced in \cref{figure_vis_results} ((a) Tropical Storm Bill, (b) a cold front, (c) a hailstorm).
    We compare three generative models, each using an ensemble size of 16.
    The ensemble means are denoted by the solid lines, with the translucent bands representing a confidence interval of one standard deviation.
    Compared to baseline models, WassDiff produces better-calibrated precipitation estimates with tighter confidence intervals, especially for the extremes (shown in red bounding boxes).}
	\label{fig_qq}
\end{figure*}

\begin{figure*}[ht]
	\centering
	\includegraphics[width=1\linewidth]{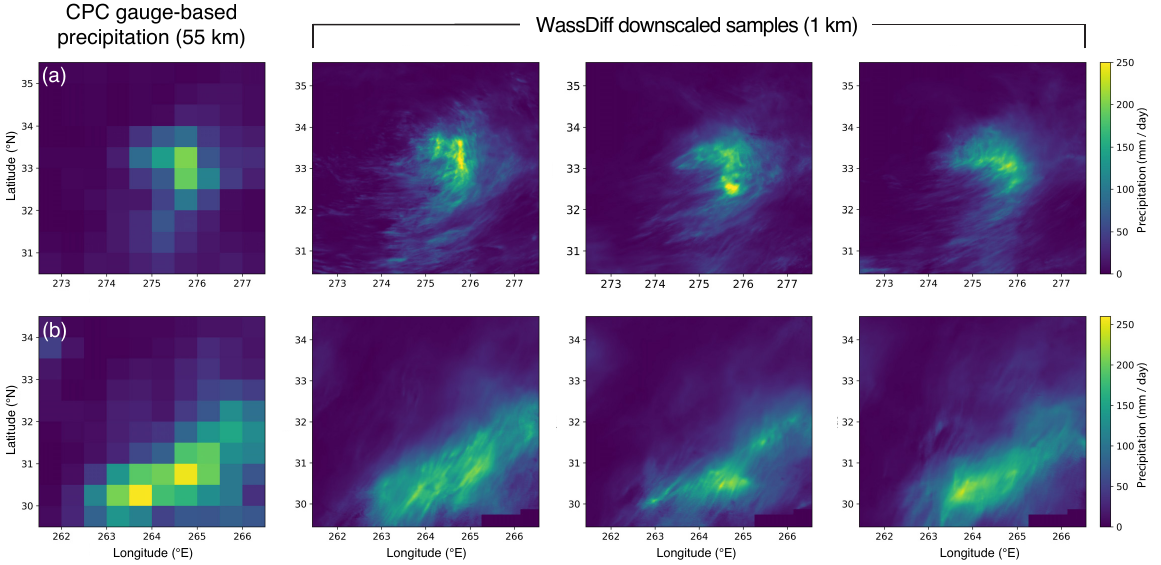}
	\caption{\textbf{High-resolution (\SI{1}{\kilo\meter}) reconstruction of historical extreme weather events using WassDiff.}
	(a) Tropical Storm Alberto, 1994-07-05 UTC.
	(b) Southeast Texas Storm, 1994-10-17 UTC.
        Predating modern km-resolution precipitation radars, these historical extreme events are only observed in gauge-based records, whose coarse resolution obscures local precipitation dynamics. Using WassDiff to perform ensemble hindcasts, researchers can now gain additional insights on these historical events that would otherwise not be possible when relying on raw measurements alone.
        }
	\label{fig_historical_results_main}
\end{figure*}

We use quantile-quantile plots (a.k.a. Q-Q plots) to measure the calibration of ensemble forecasts across different rainfall intensity levels, with a particular emphasis on the extremes.
Fig.~\ref{fig_qq} shows Q-Q plots for the three aforementioned weather events in \cref{figure_vis_results}, comparing the 0th -- 100th percentile rainfall intensity in the prediction ensemble versus the target. 
A perfectly calibrated model produces samples whose rainfall intensity estimates exactly match the target across all percentiles, which would register an expected trend line denoted by the black dashed lines in Fig.~\ref{fig_qq}.
We compared WassDiff against two baseline diffusion models (both trained with standard score-matching objective~\citep{song2020scorebased}).
We perform probabilistic inference with 16-member ensembles across all three models, where solid lines indicate ensemble means and translucent bands denote $\pm 1$ standard deviation from the ensemble mean.

Both baselines suffer from miscalibration.
SBDM exhibits a strong negative bias for almost all quantiles with wide confidence intervals (neither accurate nor confident).
CorrDiff has a comparably tighter confidence interval, but unfortunately has certain biases in different quantiles, with a significant negative bias around the third quartile and a positive bias at the extremes.
CorrDiff's biased predictions with tight forecast intervals lead to observations falling outside of the prediction interval (dashed line not intersecting with translucent bands). 

Comparatively, WassDiff is better-calibrated across the entire range of precipitation values.
The good agreement between the forecast and target suggests WassDiff captures the correct rainfall distribution, including the extremes.
Additionally, the confidence intervals for WassDiff are also tighter than SBDM across all quantiles and tighter than CorrDiff for the extremes, reflecting higher forecast precision.
This combination of accurate ensemble mean with a tight confidence interval highlights WassDiff’s robustness as a probabilistic rainfall predictor across both common and rare events.
WassDiff's better prediction for rainfall extremes is consistent with its superior MPPE and HRRE scores in \cref{tab:model_comparison}.

\subsection{Reconstruction of historical events} \label{sec_historical_events}

The input data sources (CPC~\citep{xie2007gaugebased} and ERA5~\citep{era5data}) used by WassDiff are globally available and have decades of archived data (1979 to present). 
Therefore, WassDiff can serve as a tool for researchers to reconstruct arbitrary historical events dating back to 1979, for which there are no kilometer-scale precipitation products (like MRMS) due to the limitations of instruments at the time. 
\cref{fig_historical_results_main} visualizes CPC gauge-based precipitation for two of the most severe precipitation events in CONUS history since 1979.
At \SI{55}{\kilo\meter} resolution, gauge-based precipitation only resolves limited details of the precipitation event.
In particular, extreme precipitation values are not reliable in gauge-based products due to dilution effects (averaging over \SI{55}{\kilo\meter}) and measurement errors such as wind-induced undercatch \citep{pollock2018quantifying}.
In this context, WassDiff significantly improves our understanding of historical extreme precipitation events, by providing \SI{1}{\kilo\meter} high-resolution reconstructions, as seen in \cref{fig_historical_results_main}.
We show multiple downscaled samples produced by WassDiff for each event to illustrate ensemble inference, which can be used to measure uncertainty in the reconstructions.
High-resolution reconstructions with uncertainty quantification can ultimately improve risk assessments and decision-making in climate and hydrological applications.

%% file: sections/discussion.tex
\section{Discussion and conclusion} \label{sec_discussion}

We have introduced WassDiff, a novel generative model that downscales CPC gauge-based precipitation products and ERA5 reanalysis data to generate \SI{1}{\kilo\meter}-resolution precipitation estimates.
To effectively recover extreme precipitation signals, we propose to train our score-based diffusion model with an additional distribution-matching regularization term, computed between predictions and targets.
This regularization term is calculated via Wasserstein distance, a metric that effectively captures distribution shifts for precipitation data, including distribution tails corresponding to the extremes.
Using quantile-quantile analysis, we have shown that WassDiff predictions are well-calibrated to true radar targets, particularly for peak rainfall intensity. 
Through case studies, we show that WassDiff can downscale extreme weather phenomena---such as tropical storms, cold fronts, and hailstorms---producing appropriate spatial patterns while recovering the extremes.
Comprehensive benchmarks show that WassDiff achieves state-of-the-art downscaling skills, surpassing existing generative and other ML-based downscaling baselines.

\begin{figure}[h]
    \centering
    \includegraphics[width=\linewidth]{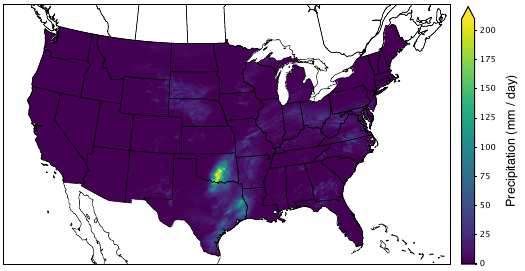}
    \caption{\textbf{Demonstration of continental-scale downscaling.}
    Using an inference-time technique called tiled diffusion~\citep{jimenez2023mixture}, WassDiff is capable of reconstructing large-scale coherent structures, such as Tropical Storm Bill (2015) in this figure, at \SI{1}{\kilo\meter} resolution across the entire contiguous United States. This supports kilometer-resolution hydrologic modeling and risk assessment across continental domains.
    }
    \label{fig_conus_gen}
\end{figure}

Downscaling via WassDiff enables the generation of extensive km-resolution precipitation datasets from decades of readily available global gauge records and reanalysis products.
Within the contiguous United States (CONUS), WassDiff expands access to \SI{1}{\kilo\meter}-resolution precipitation products from several years to several decades (1979 -- present), while also enabling generation of outputs at continental scales (as shown in \cref{fig_conus_gen}) through an inference-time technique called tiled-diffusion~\citep{jimenez2023mixture} (discussed in Supplementary).

This high-resolution and long-duration data can be used both to improve scientific understanding and to support application-driven research.
For example, kilometer-scale resolution can capture the fine-scale spatial structure and intensity of extreme rainfall, yielding deeper insights into convective storm behavior and statistics.
Similarly,  long-record high-resolution rainfall records can be used to improve intensity-duration-frequency curves, widely used in stormwater management, to understand the drivers of past floods, or to build storm catalogs for synthetic event generation.

In a broader context, this work represents a step forward in advancing the spatiotemporal resolution of precipitation records.
While we made an effort to bridge the gap between gauge and advanced radar precipitation measurements in the domain of spatial resolution, the subsequent and equally crucial step is to reduce the temporal resolution, going from daily to hourly and even minute scales.

In this study, we trained and evaluated our model on CONUS.
Since the input data source for WassDiff is globally available, in theory, WassDiff can be directly deployed to generate km-resolution precipitation outside of CONUS, potentially on other continents.
However, we do acknowledge that CONUS is a region with a relatively high gauge density.
While WassDiff is trained on a diverse set of weather phenomena ($5.5$ years of CONUS data), downscaling weather events well outside of the training distribution may also lead to less accurate estimates.
As such, deploying WassDiff outside of CONUS requires further evaluation.

Most of our efforts in this work have focused on enhancing and evaluating downscaling performance, while inference speed remains an area for further optimization, currently taking 20 seconds on a single A100 GPU per \SI{512}{\kilo\meter\squared} target. 
Although the current implementation of WassDiff supports on-demand downscaling of selected targets, further increasing its runtime efficiency would enable the open release of pre-computed, decade-long, kilometer-resolution archives readily available to the community. 
WassDiff's inference speed can be improved using several well-established techniques developed for diffusion models, including latent diffusion \citep{rombach2022highresolution}, memory-efficient attention implementations \citep{xFormers2022}, reduced or mixed precision arithmetics, and model distillation \citep{salimans2022progressive, zheng2023fast}.

While this work specifically addresses precipitation downscaling, the principle of Wasserstein Regularized Diffusion may be applicable to a broader range of inverse problems (estimating original signals from corrupted observations) in Earth sciences and other domains, particularly where the calibration of intensity estimates is critical. 
Other foreseeable applications include remote temperature retrieval \citep{liu2024island}, flood simulation \citep{farnham2018regional}, and beyond.

%% file: sections/s_wasserstein.tex
\section{Distribution-matching metrics for precipitation} \label{appendix_distance_metrics}

\begin{figure}[h]
    \centering
    \includegraphics[width=\linewidth]{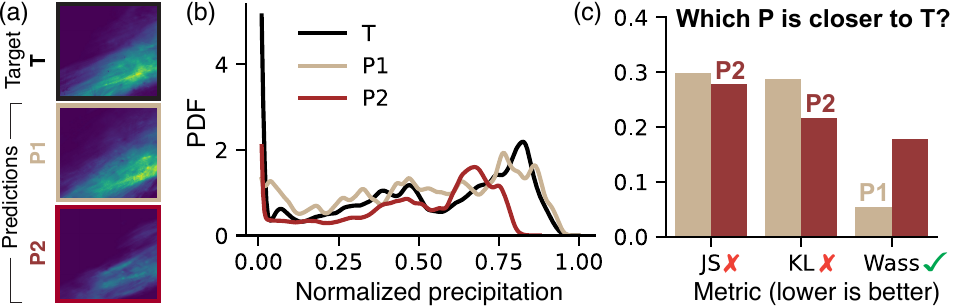}
    \caption{\textbf{Ideal distance metric for precipitation data.} (a) P1 and P2 are predictions for target T. (b) Based on their probability density functions (PDFs), P1 is a better prediction than P2. (c) However, only Wasserstein distance correctly identifies P1 as the better prediction, while KL and JS divergence incorrectly favor P2.}
    \label{fig_distance_metric}
\end{figure}

We seek a distance metric capable of quantifying the discrepancy between the sample and target distributions during the denoising process. Precipitation data is characterized by complex, heavy-tailed distributions; it is imperative to employ a metric that is sensitive to deviations in the extreme tails. 
Conventional distance measures are inadequate in this context because they predominantly capture differences around the central modes rather than penalizing errors in the tail regions.
In \cref{fig_distance_metric}, we compare Wasserstein distance with Jensen–Shannon (JS) divergence \citep{lin1991} and Kullback–Leibler (KL) divergence \citep{kullback1951}. 
Both JS and KL divergences are largely influenced by discrepancies in the modal regions of the distributions, which may lead to a misidentification of the optimal prediction (as observed with prediction P2). 
In contrast, the Wasserstein distance evaluates the cost of optimally transporting probability mass between distributions, thereby capturing differences across the entire support, including the heavy tails. 
This property renders it particularly well-suited for applications involving complex data distributions, such as those encountered in precipitation analysis.

\section{Algorithmic Details on Wasserstein Distance Regularization (WDR)} \label{appendix_sliced_wd}

At each denoising step during training, we calculate the Wasserstein distance between two distributions: generated samples (at time step $t$) and ground truth.
Given two distributions $\mathbb{P}_a$ and  $\mathbb{P}_b$, the 1D Wasserstein distance is given by:
\begin{equation} \label{eq_emd}
	W(\mathbb{P}_a, \mathbb{P}_b) = \underset{\gamma \sim \Pi(\mathbb{P}_a, \mathbb{P}_b)}{\inf} \mathbb{E}_{(k, l) \sim \gamma} \big [ \| k  - l \|| \big],
\end{equation}
\noindent where $\Pi(\mathbb{P}_a, \mathbb{P}_b)$ denotes the set of all distributions $\gamma (k, l)$ whose marginals are $\mathbb{P}_a$ and $\mathbb{P}_b$.
Since gridded precipitation data lie in high-dimensional space, we approximate this high-dimensional Wasserstein distance using sliced Wasserstein distance \citep{bonneel2015sliced}.
This is obtained by projecting high-dimensional vectors to a set of random 1D sub-planes, and then computing the average projected 1D Wasserstein distances. 

Consider a distribution of samples and targets, each with shape $[m, 1, h, w]$, where $m, h, w$ refer to the number of samples (in this case, size of a minibatch), height, and width of the images, respectively.
We first vectorize the two distributions to obtain matrices $\bv{A}, \bv{B}\in \mathbb{R}^{m, d}$, where $d := h \times w$. We computed the sliced Wasserstein distance, $W^S(\bv{A}, \bv{B})$ as follows:

\begin{algorithm}
\caption{Sliced Wasserstein distance $W^S(\bv{A}, \bv{B})$}
\label{alg:sliced_wasserstein}
\begin{algorithmic}
\Require $\bv{A}, \bv{B} \in \mathbb{R}^{m, d}, N > 0$
\For{$i = 1$ to $n$}
    \State $\mathbf{v} \sim \text{Uniform}(\mathbb{S}^{d-1})$
    \State $\bv{a}_i \gets \bv{A} \cdot \mathbf{v}$
    \State $\bv{b}_i \gets \bv{B} \cdot \mathbf{v}$
\EndFor
\State \Return $\frac{1}{N} \sum_{i}^{N} W_1(\bv{a}_i, \bv{b}_i)$
\end{algorithmic}
\end{algorithm}

\FloatBarrier

\noindent where $\mathbb{S}^{d-1}$ is a unit sphere in $\mathbb{R}^d$, and $\mathbf{v} \sim \text{Uniform}(\mathbb{S}^{n-1})$ is a random projection vector on $\mathbb{R}^n$.
$W_1(\bv{a}_i, \bv{b}_i)$ is the 1D Wasserstein distance between $\bv{a}_i$ and $\bv{b}_i$, which is computed by the area between the two marginal cumulative distribution functions (CDFs) between $\bv{a}_i$ and $\bv{b}_i$ \citep{deangelis2021why}.
In our implementation, we choose the number of random projections $N = 100$. 

\FloatBarrier

%% file: sections/s_setup.tex
\section{Experimental setup} \label{appendix_setup}

\medskip \noindent \textbf{Datasets.}
\smallskip \noindent \underline{\textit{CPC Unified Precipitation.}}
The National Oceanic and Atmospheric Administration (NOAA) Climate Prediction Center (CPC) provides a gauge-based analysis of daily precipitation products constructed on a $0.5^{\circ}$ latitude-longitude grid (approximately \SI{55}{\kilo\metre} resolution) over the entire Earth from 1978 to present \citep{xie2007gaugebased}.
In addition, we also obtain gauge network density from CPC, which describes the number of gauges per $0.25^{\circ} \times 0.25^{\circ}$ grid cell used for each daily observation.

\smallskip \noindent \underline{\textit{ERA5 Reanalysis Products.}}
The European Centre for Medium-Range Weather Forecasts (ECMWF) fifth-generation atmospheric reanalysis product (ERA5) \citep{era5data} provides global hourly estimates of atmospheric, land, and oceanic climate variables, available from 1940 to present. 
ERA5 data covers the Earth on a  $0.25^{\circ}$ latitude-longitude grid (approximately \SI{31}{\kilo\metre} resolution) and resolves the atmosphere using $137$ levels from the surface up to the height of \SI{80}{\kilo\metre}.
We use a small subset of six ERA5 variables that strongly impact precipitation: \SI{2}{\meter} temperature (\SI{}{\kelvin}), geopotential (at Earth's surface, i.e., orography) (\SI{}{\meter\squared\per\second\squared}), $\mu$ and $\nu$ components of wind (\SI{}{\metre\per\second}) at \SI{500}{\hecto\pascal}, and vertical integral of northward and eastward water vapor flux (\SI{}{\kilogram\per\meter\per\second}).

\smallskip \noindent \underline{\textit{MRMS Precipitation.}}
The Multi-Radar Multi-Sensor (MRMS) \citep{zhang2016multiradar} system was developed by NOAA's National Centers for Environmental Prediction to produce severe weather, transportation, and precipitation products.
MRMS integrates about 180 operational radars across CONUS and southern Canada along with 7000 hourly gauge, atmospheric, environmental, and climatological data to produce precipitation estimates at $0.01^{\circ}$ (\SI{\approx1}{\kilo\meter}) spatial resolution with a 2 min update cycle, which we convert to daily precipitation.
We consider MRMS daily aggregates as high-resolution ground truth observations. 



\medskip \noindent \textbf{Data processing and normalization.} 
Our dataset contains a large corpus of precipitation events sampled in the CONUS region from Sept 4, 2016 -- Dec 31, 2021. 
We perform a $80/20$ split, with training set spanning Sept 4, 2016 -- Dec 31, 2021, and validation spanning May 8, 2015 -- Sept 3, 2016.
All train and validation samples have dimension $256^2~$\SI{}{\kilo\meter}, while test samples are $512^2~$\SI{}{\kilo\meter}.
We implement random crop selection that is heavily skewed towards positive rainfall regions.
We normalize precipitation data (CPC and MRMS) via a zero-preserving log transform, $\mathbf{\tilde{y}}_p = \text{log}(\mathbf{y}_p+1) / c_p$, where $\mathbf{y}_p$ is raw precipitation data, $\mathbf{\tilde{y}}_p$ is normalized data, and $c_p$ = 5 is a precipitation scaling constant, which roughly maps MRMS to $[0, 1]$.
Meanwhile, we linearly rescale all other scalar data (e.g.,  temperature) to [0, 1], and all vector data (e.g., wind) to [-1, 1] to preserve their directional component. 

\begin{table*}[!htb]
\centering
\caption{\textbf{Summary of ML downscaling models considered in this study.}
This table presents the six downscaling models evaluated in our quantitative comparison (\cref{tab:model_comparison}). 
All models are trained from scratch on our dataset for $N_\text{iter}$ iterations.
The table also specifies the architecture and the number of parameters ($N_\text{params}$) for each model.
WDR stands for our proposed Wasserstein Distance Regularization.
}
\label{table_baseline_details}
\begin{tabular}{@{}lllll@{}}
\toprule
Model name   & Citation                                                            & Architecture                      & $N_\text{params}$ & $N_\text{iter}$ \\ \midrule
CPC-Int      &  --                                                                   & Bilinear interpolation            & None              & None            \\
CNN          & \citet{veillette2020sevir} (2020)                        & UNet                              & 7.4 M             & 120 K                \\
CorrectorGAN & \citet{price2022increasing} (2022)                        & GAN                               & 12.1 M            & 140 K           \\
CorrDiff     & \citet{mardani2025residual} (2025)                       & UNet + diffusion (score-matching) & 157 M             & 240 K           \\
SBDM         & Ours, derived from \citet{song2020scorebased} (2020) & Diffusion (score-matching)        & 61.4 M            & 200 K           \\
\textbf{WassDiff}     & \textbf{This proposed work}                                                   & Diffusion (score-matching + WDR)  & 61.4 M            & 120 K           \\ \bottomrule
\\
\end{tabular}
\end{table*}

\medskip \noindent \textbf{Implementation of training and inference.}
WassDiff is developed primarily based upon the score-based diffusion model NCSN++~\citep{song2020scorebased}. 
We used a batch size of 12 and trained for $120$~K iterations, using an exponential moving average (EMA) rate of $0.999$.
We trained on Nvidia A100 GPU and a 32-core Intel Xeon Platinum 8362 CPU with $1$ TB of DRAM. We used the Predictor-Corrector (PC) sampling scheme~\citep{song2020scorebased} for both models, discretized at $1000$ steps with the reserve diffusion predictor, one Langevin step per predictor update, and a signal-to-noise ratio of $0.16$. 
The sampling throughput is around 20 seconds per \SI{512}{\kilo\meter\squared} sample on a single Nvidia A100 GPU, using a batch size of 12.

%% file: sections/s_baselines.tex
\section{Additional details on baseline models} \label{appendix_baselines}

The baseline models used in this study are originally designed to downscale different sets of climate variables, with different resolution ratios, and in different regions. 
For fairness, we re-train all models from scratch for $N_\text{iter}$ iterations (except for CPC-Int with no trainable parameters) on our dataset, summarized in \cref{table_baseline_details}
We specify the architecture and the number of parameters ($N_\text{params}$) for each model.
All models are trained for at least $120$~K iterations, with WassDiff receiving the fewest number of steps while still providing state-of-the-art performance.
We closely follow the original training paradigm for baselines, some of which therefore accumulated additional training steps.
CorrectorGAN's \citep{price2022increasing} three-staged training accumulated $20$~K steps prior to $120$~K steps of adversarial training.
CorrDiff has a two-stage training procedure \citet{mardani2025residual}: $120$~K iterations for regression, followed by  $120$~K iterations for diffusion.
CorrDiff additionally has higher model capacity; its $N_\text{params}$ is almost two times higher than WassDiff.
Finally, we observed that SBDM failed to converge at $120$~K steps, and therefore trained SBDM for an additional $100$~K for fairness.

%% file: sections/s_metrics.tex
\section{Verification metrics} \label{appendix_metrics}

Consider a total of $N$ $\bv{x}$ observations and $\bar{\bv{x}}$ predictions (samples), where each pixel represents the daily precipitation rainfall with unit $\text{mm/day}$. We start with \textit{Deterministic metrics}:

\noindent \smallskip \textbf{Mean absolute error (MAE)} is defined as $\frac{1}{N}\sum_{i=1}^N	 |\bar{\bv{x}}_i - \bv{x}_i|$,
where as \textbf{bias} is given by $\frac{1}{N} \sum_{i=1}^N (\bar{\bv{x}}_i - \bv{x}_i)$.
Positive and negative bias correspond to over- and under-estimation; 
A perfectly calibrated model has zero bias.

\begin{figure*}[ht]
	\centering
	\includegraphics[width=1.0\linewidth]{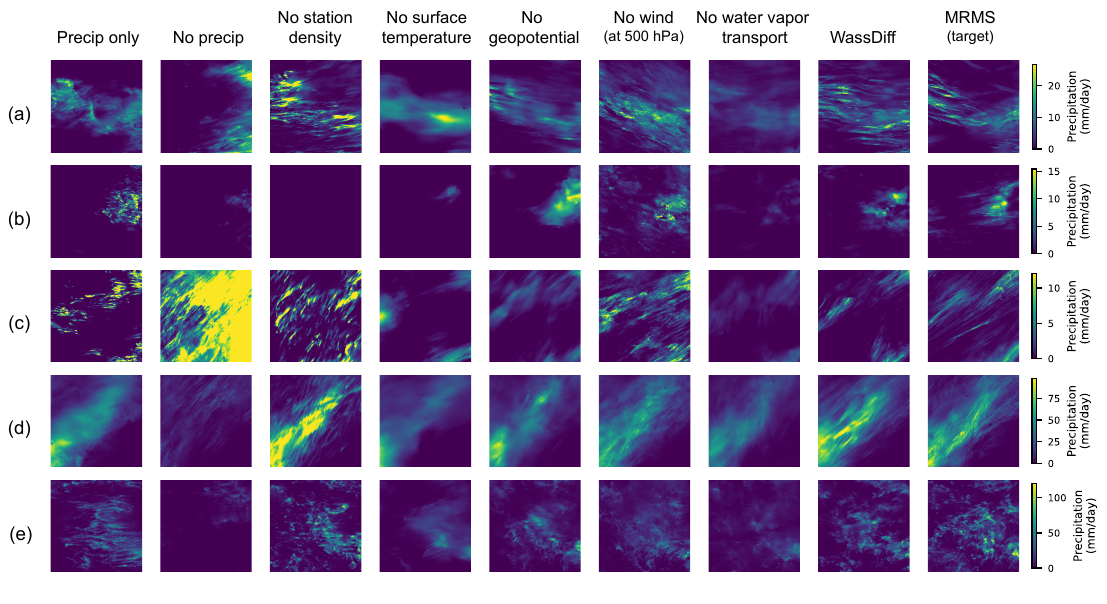}
	\caption{\textbf{Visual comparison of ablation models.}
        This figure shows samples generated by ablation models in Table~\ref{tab:ablation_comparison}.}
	\label{fig_additional_results_ablation}
\end{figure*}

\medskip \noindent \textbf{Critical Success Index (CSI)} provides a single summary of binary classification performance that rewards both precision and recall.
It evaluates whether or not rainfall exceeds a certain threshold $t$ (10~\text{mm/day} in this paper), and is defined as
\begin{equation} \label{eq_csi}
	\text{CSI} = \frac{\text{TP}}{\text{TP} + \text{FP} + \text{FN}} ,
\end{equation}
where TP, FP, and FN stand for true positive ($\bv{x}_i \ge t,~ \bar{\bv{x}} \ge t$), false positive ($\bv{x}_i \ge t,~ \bar{\bv{x}} < t$), and false negative ($\bv{x}_i < t,~ \bar{\bv{x}} \ge t$), where $i$ denote pixel location.
CSI is a monotonic transformation of $f_1$ score, where $\text{CSI} = f_1 / (2 - f_1)$.
In this study, we use the averaged pooled CSI with a pooling scale of \SI{16}{\kilo\meter}.
Pooled CSI relaxes the locality constraint and evaluates if the model gets the ``big picture'' correct.

\medskip \noindent \textbf{Learned Perceptual Image Patch Similarity (LPIPS)}~\citep{zhang2018unreasonable} assesses the perceptual similarity between images.
LPIPS evaluates similarity based on features extracted by deep neural networks, reflecting more closely on how humans perceive visual similarity.
Unlike traditional metrics that assess pixel-level accuracy, LPIPS better captures visual patterns and structures in the image that are likely relevant for interpreting meteorological conditions.

\smallskip Following \citet{chen2022rainnet} (RainNet), we use two metrics that evaluate model performance only in heavy or extreme rainfall regions: heavy rain region error (HRRE) to evaluate performance in \textit{heavy rainfall regions}, and mesoscale peak precipitation error (MPPE) to evaluate performance in \textit{extreme rainfall regions}.

\medskip \noindent \textbf{Heavy Rain Region Error (HRRE)} measures the difference in the number of \textit{threshold exceedances} (i.e., pixel count) between the sample and observation.
Following \citet{chen2022rainnet}, we define heavy rainfall regions $\mathbb{H}$ as areas (pixels) where rainfall exceeds $56$~mm/day.
This metric is defined as 

\begin{equation} \label{eq_hrre}
	\text{HRRE} = ||\mathbb{H}_{\bv{x}}| - |\mathbb{H}_{\bar{\bv{x}}}| |,
\end{equation}
\noindent where $|\mathbb{H}|$ denotes the cardinality (pixel count) of set $\mathbb{H}$.
HRRE is comparable to R20mm in CLIMDEX \citep{zhang2004rclimdex}. 

\medskip \noindent \textbf{Mesoscale Peak Precipitation Error (MPPE)} evaluates a model's ability to capture mesoscale peak precipitation. Specifically, it quantifies the error at the top $1/1000$ quantile of precipitation values between the sample and observation. A low MPPE score indicates that the sample accurately represents extreme precipitation values, irrespective of their spatial localization.
This metric is comparable to R99p in CLIMDEX \citep{zhang2004rclimdex} by definition.

\smallskip We additionally use one \textit{Ensemble metric}:

\smallskip \textbf{Continuous Ranked Probability Score (CRPS)}\citep{gneiting2007strictly} is a proper scoring rule for univariate distributions.
We use CRPS to evaluate the per-grid-cell marginals of a model's predictive distribution against observations.
CRPS is defined as
\begin{equation}
	\text{CRPS}(F, \bv{x}) = \int_{-\infty}^\infty \big(F(\bar{\bv{x}}) - \mathbf{1}\{\bar{\bv{x}} \geq \bv{x} \}\big)^2 \, \dd \bar{\bv{x}},
\end{equation}

\noindent where \( F(\bar{\bv{x}}) \) is the forecasted cumulative distribution function (CDF), \( \mathbf{1}\{\bar{\bv{x}} \geq \bv{x}\} \) is the indicator function, equal to 1 if \( \bar{\bv{x}} \geq \bv{x} \) and 0 otherwise, and $\bv{x}$ is observation.

\begin{table*}[ht]
\centering
\caption{
\textbf{Skill scores for ablation models on conditional inputs.}
We train ablation models with some subsets of variables removed from training and inference. This table presents test metrics with one ensemble member. The total number of conditional variables (\#) for each model is shown in parentheses.
}
\label{tab:ablation_comparison}
\begin{tabular}{@{}lccccccccc@{}}
\toprule
Access to conditional variables (\#) & MAE~$\downarrow$ & CSI~$\uparrow$ & Bias & HRRE $\downarrow$ & MPPE $\downarrow$ & LPIPS $\downarrow$ \\ \midrule
CPC precipitation only (1) & 3.21 {\scriptsize $\pm$ 2.48} & 0.26 {\scriptsize $\pm$ 0.27} & -0.31 {\scriptsize $\pm$ 1.78} & 1147 {\scriptsize $\pm$ 3266} & 19.65 {\scriptsize $\pm$ 22.36} & 0.47 {\scriptsize $\pm$ 0.14}\\
No CPC precipitation (7) & 4.51 {\scriptsize $\pm$ 4.38} & 0.06 {\scriptsize $\pm$ 0.12} & -2.19 {\scriptsize $\pm$ 4.73} & 1432 {\scriptsize $\pm$ 4190} & 27.60 {\scriptsize $\pm$ 28.37} & 0.51 {\scriptsize $\pm$ 0.15}\\
No CPC station density (7) & 3.85 {\scriptsize $\pm$ 3.40} & 0.26 {\scriptsize $\pm$ 0.26} & 0.40 {\scriptsize $\pm$ 1.76} & 1664 {\scriptsize $\pm$ 3805} & 21.08 {\scriptsize $\pm$ 25.19} & 0.46 {\scriptsize $\pm$ 0.13}\\
No surface temperature (7) & 3.03 {\scriptsize $\pm$ 2.45}& 0.27 {\scriptsize $\pm$ 0.27}& -0.85 {\scriptsize $\pm$ 1.80}& 1269 {\scriptsize $\pm$ 3572}& 21.77 {\scriptsize $\pm$ 23.49}& 0.49 {\scriptsize $\pm$ 0.17}\\
No geopotential (7) & 3.29 {\scriptsize $\pm$ 6.49} & 0.27 {\scriptsize $\pm$ 0.27} & 0.10 {\scriptsize $\pm$ 6.54} & 2112 {\scriptsize $\pm$ 14890} & 19.69 {\scriptsize $\pm$ 25.66} & 0.49 {\scriptsize $\pm$ 0.15}\\
No wind at 500 hPa ($\mu$ \& $\nu$) (6) & 3.62 {\scriptsize $\pm$ 2.85} & 0.27 {\scriptsize $\pm$ 0.27} & 1.16 {\scriptsize $\pm$ 2.38} & 1490 {\scriptsize $\pm$ 6509} & 15.98 {\scriptsize $\pm$ 19.96} & \textbf{0.44} {\scriptsize $\pm$ 0.09}\\
No water vapor transport (n \& e) (6) & \textbf{2.82} {\scriptsize $\pm$ 2.39} & 0.22 {\scriptsize $\pm$ 0.28} & -1.10 {\scriptsize $\pm$ 1.80} & 1377 {\scriptsize $\pm$ 3866} & 25.84 {\scriptsize $\pm$ 25.69} & 0.50 {\scriptsize $\pm$ 0.16}\\
\midrule
\textbf{All~[WassDiff]}~(8) & 3.11 {\scriptsize $\pm$ 2.64}& \textbf{0.28} {\scriptsize $\pm$ 0.28}& \textbf{-0.12} {\scriptsize $\pm$ 1.50}& \textbf{729} {\scriptsize $\pm$ 2286}& \textbf{12.65} {\scriptsize $\pm$ 14.68}& \textbf{0.44} {\scriptsize $\pm$ 0.13} \\
\bottomrule
\end{tabular}
\end{table*}

%% file: sections/s_ablation_study.tex
\section{Ablation study} \label{appendix_ablation}



In this section, we evaluate the contribution of each conditional input variable to overall model performance.
The full model, WassDiff,  has eight conditional variables as input (two from CPC and six from ERA5).\footnote{~Two ERA5 variables, wind at 500 hPa and water vapor transport, are vector fields containing two orthogonal components---eastward and northward. For wind, eastward and northward components are called u and v components, respectively. For visualizations in this paper, they are mapped to vector field diagrams, with color denoting the norm of each vector.} 
Here, we iteratively remove one input and train from scratch.
All other parameters follow WassDiff, including total number of training iterations, at 120 K.
We use the first $293$ samples in the test set, or $66\%$ of the total test samples.

Table~\ref{tab:ablation_comparison} shows test metrics of ablation models along with WassDiff.
WassDiff has higher skill metrics than all ablation models, except for MAE, which falls behind the model with no surface temperature and no water vapor transport. 
However, Fig.~\ref{fig_additional_results_ablation} shows that the two models with lower MAE scores produce significantly blurrier outputs, which is undesirable as they suppress extreme rainfall regions. 
Fig.~\ref{fig_additional_results_ablation} also reveals that in addition to surface temperature and water vapor transport, the removal of geopotential also leads to blurrier predictions. 
LPIPS agrees with this observation, as the three models are found in the top 4 highest (worst) LPIPS scores in Table~\ref{tab:ablation_comparison}. Removal of station density leads to a drop in skill metrics across the board, and we observe that the difference is more pronounced for spatially isolated rainfall events (see  Fig.~\ref{fig_additional_results_ablation}(b) and also (c)).
The model without wind at 500 hPa produces samples closest to WassDiff in terms of texture similarity (achieving the same LPIPS), but some of its metrics fall significantly behind WassDiff.
Specifically, the observation that the model without wind has significantly worse MAE but comparable CSI and MPPE suggests that wind at 500 hPa may help with better localization of extreme rainfall events.

WassDiff appears to heavily rely on both CPC precipitation and station density, as seen in Table~\ref{tab:ablation_comparison}. 
The model without CPC precipitation as input (second row) is trained to predict rainfall solely from other ERA5 climate variables. Its skill metrics are the worst among all ablation models, as expected. 
Fig.~\ref{fig_additional_results_ablation} suggests that the model without access to rainfall as input has some, but very limited, ability to predict rainfall.
Lastly, we also include a model trained without any ERA5 variables or CPC station density (first row).
Using only CPC precipitation as input, this model fails to generate textures that are consistent with MRMS targets, but in terms of accuracy in rainfall value prediction, it is comparable to other models.

\begin{figure*}[htb]
	\centering
	\includegraphics[width=0.95\linewidth]{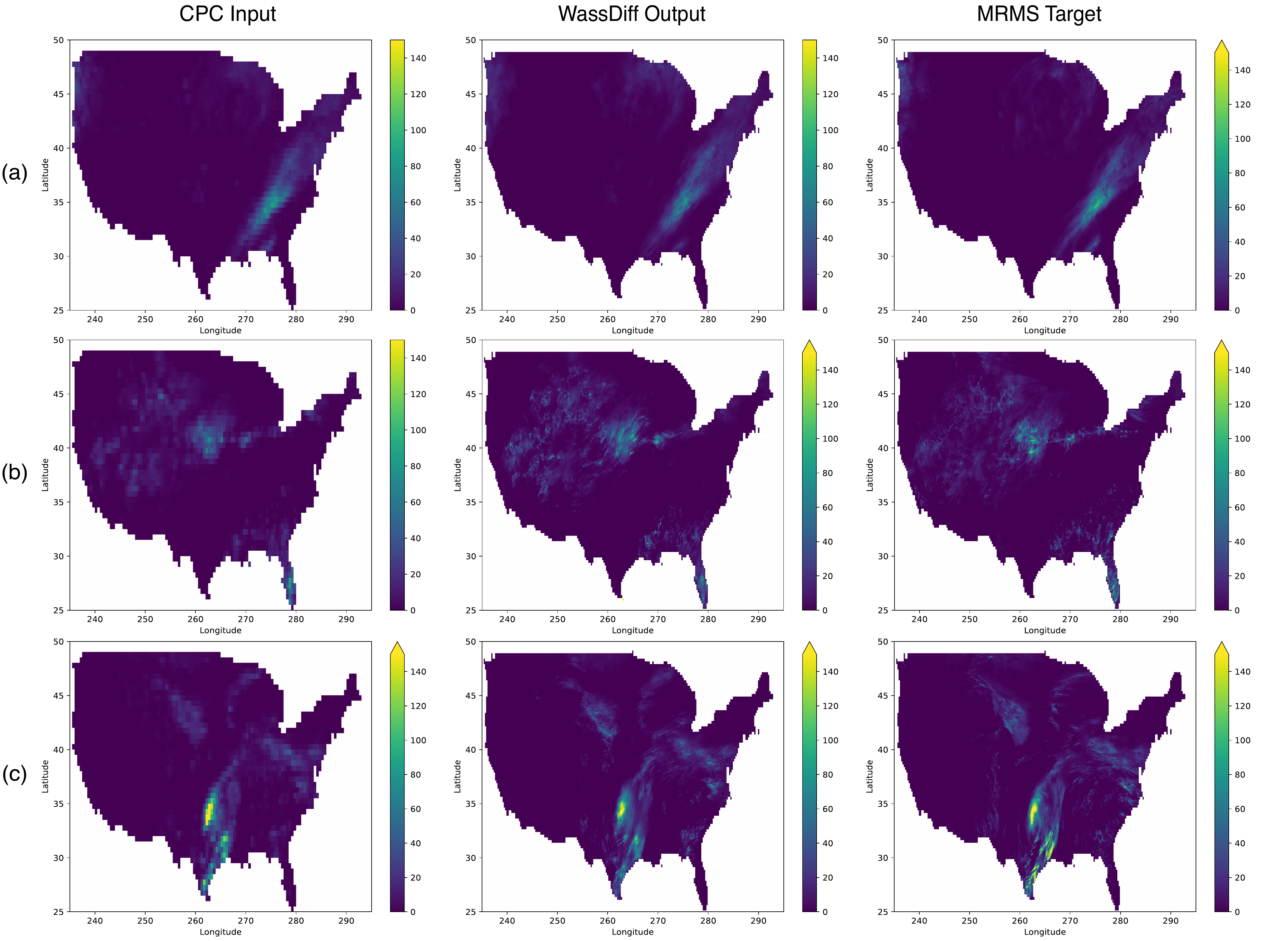}
	\caption{\textbf{Demonstration of continental-scale km-resolution downscaling.} We show \SI{1}{\kilo\meter} precipitation outputs generated by WassDiff using tiled diffusion.
    This figure showcases downscaling at the continental scale, spanning across the entire CONUS. Each figure panel shows precipitation (\SI{}{\milli\meter\per\day}) corresponding to distinct meteorological events: (a) A cold front, 2015-12-02 UTC. (b) A hailstorm, 2015-06-11 UTC. (c) Storm Bill, 2015-06-08 UTC.}
	\label{fig_tiled}
\end{figure*}

%% file: sections/s_tiled_diffusion.tex
\section{Tiled Diffusion Implementation Details} \label{appendix_tiled_diffusion}

The finite video random access memory (VRAM) capacity in GPUs naturally limits deep learning models from generating arbitrarily large images. 
In this case, each denoising step of WassDiff is carried out by a UNet, which contains attention layers, leading to a VRAM consumption that scales super-linearly with respect to input dimension.
To overcome this limitation, we implement a tiled diffusion approach following \citet{jimenez2023mixture}.
This method partitions an input image into smaller tiles (i.e., patches) such that denoising UNet can be applied to individual patches rather than the entire image, thereby lowering the total VRAM consumption.
We merge neighboring patches at each denoising step, allowing us to form a seamless, full image output.

Merging neighboring patches requires appropriately assigning weights to different pixels in the image. 
Here we use Gaussian kernels for weight assignment, applied to each patch during the denoising process.
Such a process is governed by two hyperparameters: patch size (the dimension of each patch) and stride (the spacing between adjacent patches). 
In this work, the patch size is set to 256 pixels; the stride is defined as 192 pixels.
The Gaussian weighting emphasizes the central region of the patch while tapering off towards the boundaries, ensuring that overlapping patches blend smoothly and minimize reconstruction artifacts during merging. 
The merged image is used as input to the next denoising step. 
We repeat until we approach $t=0$, where we obtain a denoised sample, such as the result shown in \cref{fig_tiled}.


\FloatBarrier

%% file: MAIN.bbl
\begin{thebibliography}{65}
\providecommand{\natexlab}[1]{#1}
\providecommand{\url}[1]{\texttt{#1}}
\expandafter\ifx\csname urlstyle\endcsname\relax
  \providecommand{\doi}[1]{doi: #1}\else
  \providecommand{\doi}{doi: \begingroup \urlstyle{rm}\Url}\fi

\bibitem[Wright et~al.(2019)Wright, Bosma, and Lopez-Cantu]{wright2019us}
Daniel~B Wright, Christopher~D Bosma, and Tania Lopez-Cantu.
\newblock Us hydrologic design standards insufficient due to large increases in frequency of rainfall extremes.
\newblock \emph{Geophysical Research Letters}, 46\penalty0 (14):\penalty0 8144--8153, 2019.

\bibitem[Calvin et~al.(2023)Calvin, Dasgupta, Krinner, Mukherji, Thorne, Trisos, et~al.]{calvin2023ipcc}
Katherine Calvin, Dipak Dasgupta, Gerhard Krinner, Aditi Mukherji, Peter~W Thorne, Christopher Trisos, et~al.
\newblock {IPCC}, 2023: Climate change 2023: Synthesis report.
\newblock \emph{First. Intergovernmental Panel on Climate Change (IPCC)}, 2023.

\bibitem[Seneviratne et~al.(2021)Seneviratne, Zhang, Adnan, Badi, Dereczynski, Di~Luca, Ghosh, Iskander, Kossin, Lewis, et~al.]{seneviratne2021weather}
Sonia~I Seneviratne, Xuebin Zhang, Muhammad Adnan, Wafae Badi, Claudine Dereczynski, Alejandro Di~Luca, Subimal Ghosh, I~Iskander, James Kossin, Sophie Lewis, et~al.
\newblock \emph{Weather and climate extreme events in a changing climate (Chapter 11)}.
\newblock Cambridge University Press, 2021.

\bibitem[Skofronick-Jackson et~al.(2017)Skofronick-Jackson, Petersen, Berg, Kidd, Stocker, Kirschbaum, Kakar, Braun, Huffman, Iguchi, et~al.]{skofronick2017global}
Gail Skofronick-Jackson, Walter~A Petersen, Wesley Berg, Chris Kidd, Erich~F Stocker, Dalia~B Kirschbaum, Ramesh Kakar, Scott~A Braun, George~J Huffman, Toshio Iguchi, et~al.
\newblock The global precipitation measurement (gpm) mission for science and society.
\newblock \emph{Bulletin of the American Meteorological Society}, 98\penalty0 (8):\penalty0 1679--1695, 2017.

\bibitem[R{\"o}zer et~al.(2019)R{\"o}zer, Kreibich, Schr{\"o}ter, M{\"u}ller, Sairam, Doss-Gollin, Lall, and Merz]{rozer2019probabilistic}
Viktor R{\"o}zer, Heidi Kreibich, Kai Schr{\"o}ter, Meike M{\"u}ller, Nivedita Sairam, James Doss-Gollin, Upmanu Lall, and Bruno Merz.
\newblock Probabilistic models significantly reduce uncertainty in hurricane harvey pluvial flood loss estimates.
\newblock \emph{Earth's Future}, 7\penalty0 (4):\penalty0 384--394, 2019.

\bibitem[Sampson et~al.(2015)Sampson, Smith, Bates, Neal, Alfieri, and Freer]{sampson2015high}
Christopher~C Sampson, Andrew~M Smith, Paul~D Bates, Jeffrey~C Neal, Lorenzo Alfieri, and Jim~E Freer.
\newblock A high-resolution global flood hazard model.
\newblock \emph{Water resources research}, 51\penalty0 (9):\penalty0 7358--7381, 2015.

\bibitem[Schneider et~al.(2014)Schneider, Becker, Finger, {Meyer-Christoffer}, Ziese, and Rudolf]{schneider2014gpcc}
Udo Schneider, Andreas Becker, Peter Finger, Anja {Meyer-Christoffer}, Markus Ziese, and Bruno Rudolf.
\newblock {{GPCC}}'s new land surface precipitation climatology based on quality-controlled in situ data and its role in quantifying the global water cycle.
\newblock \emph{Theoretical and Applied Climatology}, 115\penalty0 (1):\penalty0 15--40, January 2014.
\newblock ISSN 1434-4483.
\newblock \doi{10.1007/s00704-013-0860-x}.

\bibitem[Ahmed et~al.(2021)Ahmed, Bali, Khan, Mohamed, and Sharma]{ahmed2021improved}
Sameh~S Ahmed, Rekha Bali, Hasim Khan, Hassan~Ibrahim Mohamed, and Sunil~Kumar Sharma.
\newblock Improved water resource management framework for water sustainability and security.
\newblock \emph{Environmental Research}, 201:\penalty0 111527, 2021.

\bibitem[Hwang and Lall(2024)]{hwang2024increasing}
Jeongwoo Hwang and Upmanu Lall.
\newblock Increasing dam failure risk in the usa due to compound rainfall clusters as climate changes.
\newblock \emph{npj Natural Hazards}, 1\penalty0 (1):\penalty0 27, 2024.

\bibitem[Fowler et~al.(2021)Fowler, Lenderink, Prein, Westra, Allan, Ban, Barbero, Berg, Blenkinsop, Do, et~al.]{fowler2021anthropogenic}
Hayley~J Fowler, Geert Lenderink, Andreas~F Prein, Seth Westra, Richard~P Allan, Nikolina Ban, Renaud Barbero, Peter Berg, Stephen Blenkinsop, Hong~X Do, et~al.
\newblock Anthropogenic intensification of short-duration rainfall extremes.
\newblock \emph{Nature Reviews Earth \& Environment}, 2\penalty0 (2):\penalty0 107--122, 2021.

\bibitem[Lanza and Stagi(2008)]{lanza2008certified}
LG~Lanza and L~Stagi.
\newblock Certified accuracy of rainfall data as a standard requirement in scientific investigations.
\newblock \emph{Advances in geosciences}, 16:\penalty0 43--48, 2008.

\bibitem[Xie et~al.(2007)Xie, Chen, Yang, Yatagai, Hayasaka, Fukushima, and Liu]{xie2007gaugebased}
Pingping Xie, Mingyue Chen, Song Yang, Akiyo Yatagai, Tadahiro Hayasaka, Yoshihiro Fukushima, and Changming Liu.
\newblock A {{Gauge-Based Analysis}} of {{Daily Precipitation}} over {{East Asia}}.
\newblock \emph{Journal of Hydrometeorology}, 8\penalty0 (3):\penalty0 607--626, June 2007.
\newblock ISSN 1525-7541, 1525-755X.
\newblock \doi{10.1175/JHM583.1}.

\bibitem[{Met Office}(2003)]{metoffice2003km}
{Met Office}.
\newblock 1 km {{Resolution UK Composite Rainfall Data}} from the {{Met Office Nimrod System}}, 2003.

\bibitem[Zhang et~al.(2016)Zhang, Howard, Langston, Kaney, Qi, Tang, Grams, Wang, Cocks, Martinaitis, Arthur, Cooper, Brogden, and Kitzmiller]{zhang2016multiradar}
Jian Zhang, Kenneth Howard, Carrie Langston, Brian Kaney, Youcun Qi, Lin Tang, Heather Grams, Yadong Wang, Stephen Cocks, Steven Martinaitis, Ami Arthur, Karen Cooper, Jeff Brogden, and David Kitzmiller.
\newblock Multi-{{Radar Multi-Sensor}} ({{MRMS}}) {{Quantitative Precipitation Estimation}}: {{Initial Operating Capabilities}}.
\newblock \emph{Bulletin of the American Meteorological Society}, 97\penalty0 (4):\penalty0 621--638, April 2016.
\newblock ISSN 0003-0007, 1520-0477.
\newblock \doi{10.1175/BAMS-D-14-00174.1}.

\bibitem[Dowell et~al.(2022)Dowell, Alexander, James, Weygandt, Benjamin, Manikin, Blake, Brown, Olson, Hu, et~al.]{dowell2022high}
David~C Dowell, Curtis~R Alexander, Eric~P James, Stephen~S Weygandt, Stanley~G Benjamin, Geoffrey~S Manikin, Benjamin~T Blake, John~M Brown, Joseph~B Olson, Ming Hu, et~al.
\newblock The high-resolution rapid refresh (hrrr): An hourly updating convection-allowing forecast model. part i: Motivation and system description.
\newblock \emph{Weather and Forecasting}, 37\penalty0 (8):\penalty0 1371--1395, 2022.

\bibitem[Routray et~al.(2010)Routray, Mohanty, Niyogi, Rizvi, and Osuri]{routray2010simulation}
A~Routray, UC~Mohanty, Dev Niyogi, SRH Rizvi, and Krishna~K Osuri.
\newblock Simulation of heavy rainfall events over indian monsoon region using wrf-3dvar data assimilation system.
\newblock \emph{Meteorology and atmospheric physics}, 106:\penalty0 107--125, 2010.

\bibitem[Wilby et~al.(1998)Wilby, Wigley, Conway, Jones, Hewitson, Main, and Wilks]{wilby1998statistical}
Robert~L Wilby, TML Wigley, D~Conway, PD~Jones, BC~Hewitson, J~Main, and DS~Wilks.
\newblock Statistical downscaling of general circulation model output: A comparison of methods.
\newblock \emph{Water resources research}, 34\penalty0 (11):\penalty0 2995--3008, 1998.

\bibitem[Nishant et~al.(2023)Nishant, Hobeichi, Sherwood, Abramowitz, Shao, Bishop, and Pitman]{nishant2023comparison}
Nidhi Nishant, Sanaa Hobeichi, Steven Sherwood, Gab Abramowitz, Yawen Shao, Craig Bishop, and Andy Pitman.
\newblock Comparison of a novel machine learning approach with dynamical downscaling for australian precipitation.
\newblock \emph{Environmental Research Letters}, 18\penalty0 (9):\penalty0 094006, 2023.

\bibitem[Veillette et~al.(2020)Veillette, Samsi, and Mattioli]{veillette2020sevir}
Mark Veillette, Siddharth Samsi, and Chris Mattioli.
\newblock {{SEVIR}} : {{A Storm Event Imagery Dataset}} for {{Deep Learning Applications}} in {{Radar}} and {{Satellite Meteorology}}.
\newblock In \emph{Chen2008assessing}, volume~33, pages 22009--22019. Curran Associates, Inc., 2020.

\bibitem[Rampal et~al.(2022)Rampal, Gibson, Sood, Stuart, Fauchereau, Brandolino, Noll, and Meyers]{rampal2022highresolution}
Neelesh Rampal, Peter~B. Gibson, Abha Sood, Stephen Stuart, Nicolas~C. Fauchereau, Chris Brandolino, Ben Noll, and Tristan Meyers.
\newblock High-resolution downscaling with interpretable deep learning: {{Rainfall}} extremes over {{New Zealand}}.
\newblock \emph{Weather and Climate Extremes}, 38:\penalty0 100525, December 2022.
\newblock ISSN 2212-0947.
\newblock \doi{10.1016/j.wace.2022.100525}.

\bibitem[Rodrigues et~al.(2018)Rodrigues, Oliveira, Cunha, and Netto]{rodrigues2018deepdownscale}
Eduardo~R. Rodrigues, Igor Oliveira, Renato L.~F. Cunha, and Marco A.~S. Netto.
\newblock {{DeepDownscale}}: A {{Deep Learning Strategy}} for {{High-Resolution Weather Forecast}}, August 2018.

\bibitem[Price and Rasp(2022)]{price2022increasing}
Ilan Price and Stephan Rasp.
\newblock Increasing the accuracy and resolution of precipitation forecasts using deep generative models, 2022.

\bibitem[Leinonen et~al.(2021)Leinonen, Nerini, and Berne]{leinonen2021stochastic}
Jussi Leinonen, Daniele Nerini, and Alexis Berne.
\newblock Stochastic {{Super-Resolution}} for {{Downscaling Time-Evolving Atmospheric Fields With}} a {{Generative Adversarial Network}}.
\newblock \emph{IEEE Transactions on Geoscience and Remote Sensing}, 59\penalty0 (9):\penalty0 7211--7223, September 2021.
\newblock ISSN 1558-0644.
\newblock \doi{10.1109/TGRS.2020.3032790}.

\bibitem[Harris et~al.(2022)Harris, McRae, Chantry, Dueben, and Palmer]{harris2022generative}
Lucy Harris, Andrew T.~T. McRae, Matthew Chantry, Peter~D. Dueben, and Tim~N. Palmer.
\newblock A {{Generative Deep Learning Approach}} to {{Stochastic Downscaling}} of {{Precipitation Forecasts}}.
\newblock \emph{Journal of Advances in Modeling Earth Systems}, 14\penalty0 (10):\penalty0 e2022MS003120, 2022.
\newblock ISSN 1942-2466.
\newblock \doi{10.1029/2022MS003120}.

\bibitem[Addison et~al.(2022)Addison, Kendon, Ravuri, Aitchison, and Watson]{addison2022machine}
Henry Addison, Elizabeth Kendon, Suman Ravuri, Laurence Aitchison, and Peter~AG Watson.
\newblock Machine learning emulation of a local-scale {{UK}} climate model, November 2022.

\bibitem[Mardani et~al.(2025)Mardani, Brenowitz, Cohen, Pathak, Chen, Liu, Vahdat, Nabian, Ge, Subramaniam, Kashinath, Kautz, and Pritchard]{mardani2025residual}
Morteza Mardani, Noah Brenowitz, Yair Cohen, Jaideep Pathak, Chieh-Yu Chen, Cheng-Chin Liu, Arash Vahdat, Mohammad~Amin Nabian, Tao Ge, Akshay Subramaniam, Karthik Kashinath, Jan Kautz, and Mike Pritchard.
\newblock Residual corrective diffusion modeling for km-scale atmospheric downscaling.
\newblock \emph{Communications Earth \& Environment}, 6\penalty0 (1):\penalty0 1--10, February 2025.
\newblock ISSN 2662-4435.
\newblock \doi{10.1038/s43247-025-02042-5}.

\bibitem[Bau et~al.(2019)Bau, Zhu, Wulff, Peebles, Strobelt, Zhou, and Torralba]{bau2019seeinggangenerate}
David Bau, Jun-Yan Zhu, Jonas Wulff, William Peebles, Hendrik Strobelt, Bolei Zhou, and Antonio Torralba.
\newblock Seeing what a gan cannot generate, 2019.

\bibitem[Pandey et~al.(2025)Pandey, Pathak, Xu, Mandt, Pritchard, Vahdat, and Mardani]{pandey2025heavytailed}
Kushagra Pandey, Jaideep Pathak, Yilun Xu, Stephan Mandt, Michael Pritchard, Arash Vahdat, and Morteza Mardani.
\newblock Heavy-tailed diffusion models.
\newblock In \emph{The Thirteenth International Conference on Learning Representations}, 2025.

\bibitem[{National Centers for Environmental Prediction}(2025)]{gefs_2025}
{National Centers for Environmental Prediction}.
\newblock {Global Ensemble Forecast System (GEFS)}, 2025.
\newblock Accessed: 2025-03-26.

\bibitem[{European Centre for Medium-Range Weather Forecasts (ECMWF)}(2025)]{ifs_2025}
{European Centre for Medium-Range Weather Forecasts (ECMWF)}.
\newblock \emph{Integrated Forecasting System Documentation}.
\newblock European Centre for Medium-Range Weather Forecasts, 2025.
\newblock Accessed: 2025-03-26.

\bibitem[Pollock et~al.(2018)Pollock, O'donnell, Quinn, Dutton, Black, Wilkinson, Colli, Stagnaro, Lanza, Lewis, et~al.]{pollock2018quantifying}
Michael~Deering Pollock, Greg O'donnell, Paul Quinn, Mark Dutton, Andrew Black, ME~Wilkinson, Matteo Colli, Mattia Stagnaro, LG~Lanza, Elizabeth Lewis, et~al.
\newblock Quantifying and mitigating wind-induced undercatch in rainfall measurements.
\newblock \emph{Water Resources Research}, 54\penalty0 (6):\penalty0 3863--3875, 2018.

\bibitem[Song et~al.(2020)Song, {Sohl-Dickstein}, Kingma, Kumar, Ermon, and Poole]{song2020scorebased}
Yang Song, Jascha {Sohl-Dickstein}, Diederik~P. Kingma, Abhishek Kumar, Stefano Ermon, and Ben Poole.
\newblock Score-{{Based Generative Modeling}} through {{Stochastic Differential Equations}}.
\newblock In \emph{International {{Conference}} on {{Learning Representations}}}, October 2020.

\bibitem[Villani(2009)]{villani2009optimal}
Cédric Villani.
\newblock \emph{Optimal Transport: Old and New}, volume 338 of \emph{Grundlehren der Mathematischen Wissenschaften}.
\newblock Springer-Verlag Berlin Heidelberg, 2009.
\newblock \doi{10.1007/978-3-540-71050-9}.

\bibitem[Ghil and Lucarini(2020)]{ghil2020physics}
Michael Ghil and Valerio Lucarini.
\newblock The physics of climate variability and climate change.
\newblock \emph{Reviews of Modern Physics}, 92\penalty0 (3):\penalty0 035002, July 2020.
\newblock ISSN 0034-6861, 1539-0756.
\newblock \doi{10.1103/RevModPhys.92.035002}.

\bibitem[Ravuri et~al.(2021)Ravuri, Lenc, Willson, Kangin, Lam, Mirowski, Fitzsimons, Athanassiadou, Kashem, Madge, Prudden, Mandhane, Clark, Brock, Simonyan, Hadsell, Robinson, Clancy, Arribas, and Mohamed]{ravuri2021skillful}
Suman Ravuri, Karel Lenc, Matthew Willson, Dmitry Kangin, Remi Lam, Piotr Mirowski, Megan Fitzsimons, Maria Athanassiadou, Sheleem Kashem, Sam Madge, Rachel Prudden, Amol Mandhane, Aidan Clark, Andrew Brock, Karen Simonyan, Raia Hadsell, Niall Robinson, Ellen Clancy, Alberto Arribas, and Shakir Mohamed.
\newblock Skillful {{Precipitation Nowcasting}} using {{Deep Generative Models}} of {{Radar}}.
\newblock \emph{Nature}, 597\penalty0 (7878):\penalty0 672--677, September 2021.
\newblock ISSN 0028-0836, 1476-4687.
\newblock \doi{10.1038/s41586-021-03854-z}.

\bibitem[Selz and Craig(2015)]{selz2015upscale}
Tobias Selz and George~C. Craig.
\newblock Upscale {{Error Growth}} in a {{High-Resolution Simulation}} of a {{Summertime Weather Event}} over {{Europe}}*.
\newblock \emph{Monthly Weather Review}, 143\penalty0 (3):\penalty0 813--827, March 2015.
\newblock ISSN 0027-0644, 1520-0493.
\newblock \doi{10.1175/MWR-D-14-00140.1}.

\bibitem[Vosper et~al.(2023)Vosper, Watson, Harris, McRae, {Santos-Rodriguez}, Aitchison, and Mitchell]{vosper2023deep}
Emily Vosper, Peter Watson, Lucy Harris, Andrew McRae, Raul {Santos-Rodriguez}, Laurence Aitchison, and Dann Mitchell.
\newblock Deep {{Learning}} for {{Downscaling Tropical Cyclone Rainfall}} to {{Hazard-Relevant Spatial Scales}}.
\newblock \emph{Journal of Geophysical Research: Atmospheres}, 128\penalty0 (10):\penalty0 e2022JD038163, 2023.
\newblock ISSN 2169-8996.
\newblock \doi{10.1029/2022JD038163}.

\bibitem[Xiao et~al.(2022)Xiao, Kreis, and Vahdat]{xiao2022tackling}
Zhisheng Xiao, Karsten Kreis, and Arash Vahdat.
\newblock Tackling the generative learning trilemma with denoising diffusion {GAN}s.
\newblock In \emph{International Conference on Learning Representations}, 2022.

\bibitem[Kodali et~al.(2017)Kodali, Abernethy, Hays, and Kira]{kodali2017convergence}
Naveen Kodali, Jacob Abernethy, James Hays, and Zsolt Kira.
\newblock On {{Convergence}} and {{Stability}} of {{GANs}}, December 2017.

\bibitem[Salimans et~al.(2016)Salimans, Goodfellow, Zaremba, Cheung, Radford, and Chen]{salimans2016improved}
Tim Salimans, Ian Goodfellow, Wojciech Zaremba, Vicki Cheung, Alec Radford, and Xi~Chen.
\newblock Improved techniques for training gans.
\newblock \emph{Advances in neural information processing systems}, 29, 2016.

\bibitem[Ho et~al.(2020)Ho, Jain, and Abbeel]{ho2020denoising}
Jonathan Ho, Ajay Jain, and Pieter Abbeel.
\newblock Denoising {{Diffusion Probabilistic Models}}, December 2020.

\bibitem[Dhariwal and Nichol(2021)]{dhariwal2021diffusion}
Prafulla Dhariwal and Alex Nichol.
\newblock Diffusion {{Models Beat GANs}} on {{Image Synthesis}}, June 2021.

\bibitem[Giorgi(2019)]{giorgi2019thirty}
Filippo Giorgi.
\newblock Thirty {{Years}} of {{Regional Climate Modeling}}: {{Where Are We}} and {{Where Are We Going}} next?
\newblock \emph{Journal of Geophysical Research: Atmospheres}, 124\penalty0 (11):\penalty0 5696--5723, 2019.
\newblock ISSN 2169-8996.
\newblock \doi{10.1029/2018JD030094}.

\bibitem[Hersbach et~al.(2023)Hersbach, Bell, Berrisford, Biavati, Hor{\'a}nyi, Mu{\~n}oz~Sabater, Nicolas, Peubey, Radu, Rozum, Schepers, Simmons, Soci, Dee, and Th{\'e}paut]{era5data}
H.~Hersbach, B.~Bell, P.~Berrisford, G.~Biavati, A.~Hor{\'a}nyi, J.~Mu{\~n}oz~Sabater, J.~Nicolas, C.~Peubey, R.~Radu, I.~Rozum, D.~Schepers, A.~Simmons, C.~Soci, D.~Dee, and J.-N. Th{\'e}paut.
\newblock Era5 hourly data on single levels from 1940 to present.
\newblock Copernicus Climate Change Service (C3S) Climate Data Store (CDS), 2023.
\newblock Accessed on 29-Apr-2024.

\bibitem[Back and Bretherton(2005)]{back2005relationship}
Larissa~E Back and Christopher~S Bretherton.
\newblock The relationship between wind speed and precipitation in the pacific itcz.
\newblock \emph{Journal of climate}, 18\penalty0 (20):\penalty0 4317--4328, 2005.

\bibitem[Li et~al.(2024)Li, Li, Zhao, Cao, Li, and Zhong]{li2024revealing}
Weidong Li, Tiejian Li, Jie Zhao, Yuan Cao, Zhaoxi Li, and Deyu Zhong.
\newblock Revealing circulation patterns responsible for extreme precipitation events over the hai river basin from moisture transport perspective.
\newblock \emph{Journal of Geophysical Research: Atmospheres}, 129\penalty0 (23):\penalty0 e2024JD040993, 2024.
\newblock ISSN 2169-8996.
\newblock \doi{10.1029/2024JD040993}.

\bibitem[O'Gorman(2015)]{ogorman2015precipitation}
Paul~A. O'Gorman.
\newblock Precipitation {{Extremes Under Climate Change}}.
\newblock \emph{Current Climate Change Reports}, 1\penalty0 (2):\penalty0 49--59, June 2015.
\newblock ISSN 2198-6061.
\newblock \doi{10.1007/s40641-015-0009-3}.

\bibitem[Song and Ermon(2020)]{song2020generative}
Yang Song and Stefano Ermon.
\newblock Generative {{Modeling}} by {{Estimating Gradients}} of the {{Data Distribution}}, October 2020.

\bibitem[Lin(1991)]{lin1991}
J.~Lin.
\newblock Divergence measures based on the {Shannon} entropy.
\newblock \emph{IEEE Transactions on Information Theory}, 37\penalty0 (1):\penalty0 145--151, 1991.

\bibitem[Kullback and Leibler(1951)]{kullback1951}
S.~Kullback and R.~A. Leibler.
\newblock On information and sufficiency.
\newblock \emph{The Annals of Mathematical Statistics}, 22\penalty0 (1):\penalty0 79--86, 1951.

\bibitem[Bonneel et~al.(2015)Bonneel, Rabin, Peyr{\'e}, and Pfister]{bonneel2015sliced}
Nicolas Bonneel, Julien Rabin, Gabriel Peyr{\'e}, and Hanspeter Pfister.
\newblock Sliced and {{Radon Wasserstein Barycenters}} of {{Measures}}.
\newblock \emph{Journal of Mathematical Imaging and Vision}, 1\penalty0 (51):\penalty0 22--45, 2015.
\newblock \doi{10.1007/s10851-014-0506-3}.

\bibitem[Kolouri et~al.(2019)Kolouri, Nadjahi, Simsekli, Badeau, and Rohde]{kolouri2019generalized}
Soheil Kolouri, Kimia Nadjahi, Umut Simsekli, Roland Badeau, and Gustavo Rohde.
\newblock Generalized sliced wasserstein distances.
\newblock \emph{Advances in neural information processing systems}, 32, 2019.

\bibitem[Nguyen and Ho(2024)]{nguyen2024energy}
Khai Nguyen and Nhat Ho.
\newblock Energy-based sliced wasserstein distance.
\newblock \emph{Advances in Neural Information Processing Systems}, 36, 2024.

\bibitem[Chen et~al.(2022)Chen, Feng, Liu, Ni, Lu, Tong, and Liu]{chen2022rainnet}
Xuanhong Chen, Kairui Feng, Naiyuan Liu, Bingbing Ni, Yifan Lu, Zhengyan Tong, and Ziang Liu.
\newblock {{RainNet}}: {{A Large-Scale Imagery Dataset}} and {{Benchmark}} for {{Spatial Precipitation Downscaling}}.
\newblock \emph{Advances in Neural Information Processing Systems}, 35:\penalty0 9797--9812, December 2022.

\bibitem[Zhang et~al.(2018)Zhang, Isola, Efros, Shechtman, and Wang]{zhang2018unreasonable}
Richard Zhang, Phillip Isola, Alexei~A. Efros, Eli Shechtman, and Oliver Wang.
\newblock The {{Unreasonable Effectiveness}} of {{Deep Features}} as a {{Perceptual Metric}}, April 2018.

\bibitem[Jim{\'e}nez(2023)]{jimenez2023mixture}
{\'A}lvaro~Barbero Jim{\'e}nez.
\newblock Mixture of diffusers for scene composition and high resolution image generation.
\newblock \emph{arXiv preprint arXiv:2302.02412}, 2023.

\bibitem[Rombach et~al.(2022)Rombach, Blattmann, Lorenz, Esser, and Ommer]{rombach2022highresolution}
Robin Rombach, Andreas Blattmann, Dominik Lorenz, Patrick Esser, and Bj{\"o}rn Ommer.
\newblock High-{{Resolution Image Synthesis}} with {{Latent Diffusion Models}}, April 2022.

\bibitem[Xiong et~al.(2021)Xiong, O{\u{g}}uz, Gupta, Chen, Liskovich, Levy, Yih, and Mehdad]{xFormers2022}
Wenhan Xiong, Barlas O{\u{g}}uz, Anchit Gupta, Xilun Chen, Diana Liskovich, Omer Levy, Wen-tau Yih, and Yashar Mehdad.
\newblock Simple local attentions remain competitive for long-context tasks.
\newblock \emph{arXiv preprint arXiv:2112.07210}, 2021.

\bibitem[Salimans and Ho(2022)]{salimans2022progressive}
Tim Salimans and Jonathan Ho.
\newblock Progressive distillation for fast sampling of diffusion models.
\newblock In \emph{International Conference on Learning Representations}, 2022.

\bibitem[Zheng et~al.(2023)Zheng, Nie, Vahdat, Azizzadenesheli, and Anandkumar]{zheng2023fast}
Hongkai Zheng, Weili Nie, Arash Vahdat, Kamyar Azizzadenesheli, and Anima Anandkumar.
\newblock Fast sampling of diffusion models via operator learning.
\newblock In \emph{International Conference on Machine Learning}, pages 42390--42402. PMLR, 2023.

\bibitem[Liu et~al.(2024)Liu, Panakkal, Dee, Balakrishnan, Padgett, and Veeraraghavan]{liu2024island}
Yuhao Liu, Pranavesh Panakkal, Sylvia Dee, Guha Balakrishnan, Jamie Padgett, and Ashok Veeraraghavan.
\newblock {{ISLAND}}: {{Interpolating Land Surface Temperature}} using land cover.
\newblock \emph{Remote Sensing Applications: Society and Environment}, 36:\penalty0 101332, November 2024.
\newblock ISSN 23529385.
\newblock \doi{10.1016/j.rsase.2024.101332}.

\bibitem[Farnham et~al.(2018)Farnham, Doss-Gollin, and Lall]{farnham2018regional}
David~J Farnham, James Doss-Gollin, and Upmanu Lall.
\newblock Regional extreme precipitation events: Robust inference from credibly simulated gcm variables.
\newblock \emph{Water Resources Research}, 54\penalty0 (6):\penalty0 3809--3824, 2018.

\bibitem[De~Angelis and Gray(2021)]{deangelis2021why}
Marco De~Angelis and Ander Gray.
\newblock Why the 1-{{Wasserstein}} distance is the area between the two marginal {{CDFs}}, November 2021.

\bibitem[Zhang et~al.(2004)Zhang, Yang, et~al.]{zhang2004rclimdex}
Xuebin Zhang, Feng Yang, et~al.
\newblock Rclimdex (1.0) user manual.
\newblock \emph{Climate Research Branch Environment Canada}, 22:\penalty0 13--14, 2004.

\bibitem[Gneiting and Raftery(2007)]{gneiting2007strictly}
Tilmann Gneiting and Adrian~E Raftery.
\newblock Strictly {{Proper Scoring Rules}}, {{Prediction}}, and {{Estimation}}.
\newblock \emph{Journal of the American Statistical Association}, 102\penalty0 (477):\penalty0 359--378, March 2007.
\newblock ISSN 0162-1459.
\newblock \doi{10.1198/016214506000001437}.

\end{thebibliography}
